\documentclass[sn-mathphys,Numbered]{sn-jnl}% Math and Physical Sciences Reference Style
\usepackage{graphicx}%
\usepackage{multirow}%
\usepackage{amsmath,amssymb,amsfonts}%
\usepackage{amsthm}%
\usepackage{mathrsfs}%
\usepackage[title]{appendix}%
\usepackage{xcolor}%
\usepackage{textcomp}%
\usepackage{manyfoot}%
\usepackage{booktabs}%
\usepackage{algorithm}%
\usepackage{algorithmic}
\usepackage{listings}%
\usepackage{amsmath,amsfonts,bm}

\def\eqref#1{equation~\ref{#1}}
\def\1{\bm{1}}

\def\vx{{\bm{x}}}

\DeclareMathAlphabet{\mathsfit}{\encodingdefault}{\sfdefault}{m}{sl}
\SetMathAlphabet{\mathsfit}{bold}{\encodingdefault}{\sfdefault}{bx}{n}

\def\sC{{\mathbb{C}}}

\def\sL{{\mathbb{L}}}

\def\sU{{\mathbb{U}}}

\def\sW{{\mathbb{W}}}
\def\sX{{\mathbb{X}}}
\def\sY{{\mathbb{Y}}}

\newcommand{\R}{\mathbb{R}}

\theoremstyle{thmstyleone}%
\newtheorem{theorem}{Theorem}%  meant for continuous numbers
\theoremstyle{thmstyletwo}%

\theoremstyle{thmstylethree}%

\begin{document}

% \section{Submission Information Sheet}

% \newpage
\title[Article Title]{Exploiting Counter-Examples for Active Learning with Partial labels}


\author[1,2]{\fnm{Fei} \sur{Zhang}}\email{ferenas@sjtu.edu.cn}

\author[3]{\fnm{Yunjie} \sur{Ye}}\email{yejunjie4@huawei.com}

\author[4]{\fnm{Lei} \sur{Feng}}\email{lfengqaq@gmail.com}

\author[3]{\fnm{Zhongwen} \sur{Rao}}\email{raozhongwen@huawei.com}

\author[3]{\fnm{Jieming} \sur{Zhu}}\email{jiemingzhu@ieee.org}

\author[3]{\fnm{Marcus} \sur{Kalander}}\email{marcus.kalander@huawei.com}

\author[5]{\fnm{Chen} \sur{Gong}}\email{chen.gong@njust.edu.cn}

\author[3]{\fnm{Jianye} \sur{Hao}}\email{jianye.hao@tju.edu.cn}

\author*[1]{\fnm{Bo} \sur{Han}}\email{bhanml@comp.hkbu.edu.hk}
%\orgdiv{},

\affil[1]{\orgname{Shanghai Jiao Tong University}, \orgaddress{\city{Shanghai}, \postcode{200240}, \country{China}}}

\affil*[2]{\orgname{Hong Kong Baptist University}, \orgaddress{\city{Hong Kong}, \postcode{999077}, \country{China}}}

\affil[3]{\orgname{Huawei Noah’s Ark Lab}, \orgaddress{\city{Shenzhen}, \postcode{518129}, \country{China}}}

\affil[4]{\orgname{Nanyang Technological University}, \orgaddress{\city{Shenzhen}, \postcode{639798}, \country{Singapore}}}

\affil[5]{\orgname{Nanjing University of Science and Technology}, \orgaddress{\city{Nanjing}, \postcode{210094}, \country{China}}}

% \affil[6]{\orgname{Tianjin University}, \orgaddress{\city{Tianjin}, \postcode{300072}, \country{China}}}

\abstract{This paper studies a new problem, \emph{active learning with partial labels} (ALPL). In this setting, an oracle annotates the query samples with partial labels, relaxing the oracle from the demanding accurate labeling process. To address ALPL, we first build an intuitive baseline that can be seamlessly incorporated into existing AL frameworks. Though effective, this baseline is still susceptible to the \emph{overfitting}, and falls short of the representative partial-label-based samples during the query process.
Drawing inspiration from human inference in cognitive science, where accurate inferences can be explicitly derived from \emph{counter-examples} (CEs), our objective is to leverage this human-like learning pattern to tackle the \emph{overfitting} while enhancing the process of selecting representative samples in ALPL. Specifically, we construct CEs by reversing the partial labels for each instance, and then we propose a simple but effective WorseNet to directly learn from this complementary pattern. By leveraging the distribution gap between WorseNet and the predictor, this adversarial evaluation manner could enhance both the performance of the predictor itself and the sample selection process, allowing the predictor to capture more accurate patterns in the data. Experimental results on five real-world datasets and four benchmark datasets show that our proposed method achieves comprehensive improvements over ten representative AL frameworks, highlighting the superiority of WorseNet. The source code will be available at \url{https://github.com/Ferenas/APLL}.}




\maketitle

\section{Introduction}\label{sec:intro}

The community of artificial intelligence has witnessed great progress owing to deep learning, whose success heavily relies on the quality and volume of accurately annotated datasets. To ease the pressure of such costing labeling work, numerous researchers have been investigating \emph{active learning} (AL)~\cite{settles1995active}, which aims to achieve as high-performance gain as possible by labeling as few samples as possible. A popular setting in AL is pool-based AL~\cite{settles1995active}, where a fixed number of samples selected by a selector are sent to an oracle for labeling iteratively until the exhaustion of the sampling budget. Pool-based AL has a wide range of applications, including but not limited to semantic segmentation~\cite{segmentation3} and object detection~\cite{detection2}.

\begin{figure}[htbp]
\centering
\vspace{-4mm}
\includegraphics[width=0.80\linewidth]{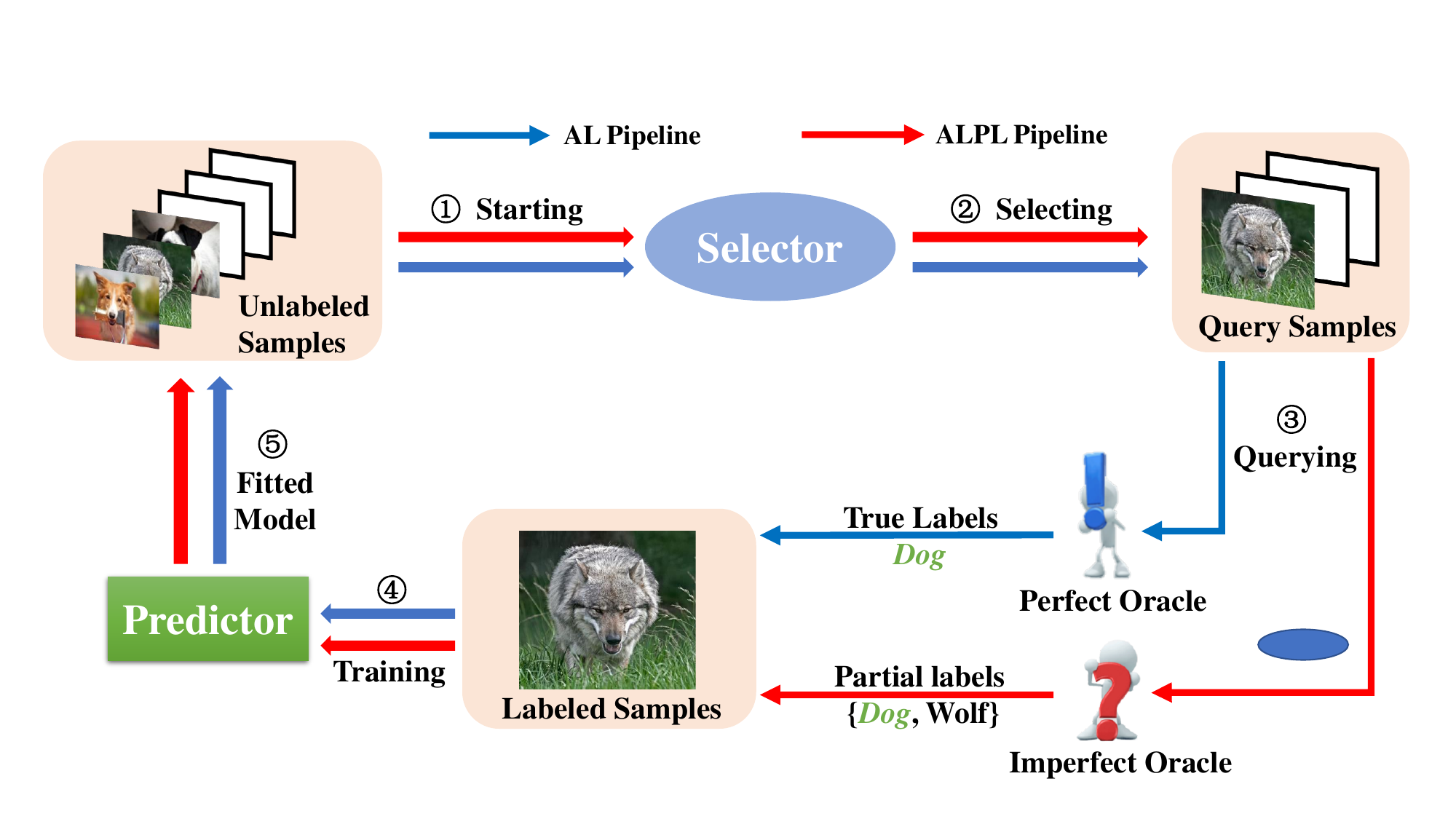}
\vspace{-4mm}
\caption{Comparison of pool-based AL (blue arrow) and our proposed ALPL framework (red arrow). The core difference between these two settings is the label form provided by the oracle.}
\label{Fig_motivation}
% \vspace{-6mm}
\end{figure}

Most existing pool-based AL frameworks~\cite{joshi2009multi,luo2013latent,lloss,batchbald,tvaal,parvaneh2022active} assume that the oracle is perfect, i.e., the oracle always provides accurate labels for selected samples. However, due to inherent label ambiguity and noise, we cannot expect such a ``perfect'' oracle to exist in real-world applications~\cite{fang2012don}. Let us consider a birdsong classification problem~\cite{briggs2012acoustic}. The
songs of different bird species are usually recorded simultaneously in one field-collected recording. Thus, it would be difficult for experts to localize each specie to the corresponding 
spectrogram simply by virtue of this recording. To apply AL in a more practical way, we turn to a new type of imperfect oracle, which would provide the selected samples with a special but prevailing form of the weak label, i.e., partial label. A partial label of an instance, essentially a set of candidate labels that includes the true label, is intuitively adaptable to various real-world tasks, including image retrieval~\cite{cour2011learning} and face recognition~\cite{zeng2013learning}. With the full potential of partial labels seen in these real-world scenarios, \emph{partial-label learning} (PLL), has naturally emerged and boomed in the community~\cite{feng2018leveraging,wang2019adaptive,wang2022pico,zhang2022exploiting}. Motivated by the industrial and academic value of PLL, we propose a new setting for AL, i.e.,~\emph{active learning with partial labels} (ALPL). Formally, ALPL is built on a pool-based AL learning problem but with only one imperfect oracle that assigns partial labels to samples. Figure~\ref{Fig_motivation} illustrates the pipelines of AL and ALPL. Compared with AL, the oracle in ALPL shall provide noise-tolerant partial labels instead of the exact true label when annotating confusing objects, highly improving the labeling efficiency while easing the annotation pressure of the oracle. 

To address ALPL, we first focus on building a group of promising baselines by adopting the RC loss~\cite{feng2020provably}, as one of the state-of-the-art milestones in PLL~\cite{lv2020progressive,wen2021leveraged,wang2022pico,zhang2022exploiting}, to train the predictor with the given partial labels from the oracle. 
% , , has been not only theoretically proved to achieve risk-consistency in PLL, but also experimentally evaluated to show competitive performance compared with various works~\cite{lv2020progressive,wen2021leveraged,wang2022pico,zhang2022exploiting}. 
%essentially a weighted cross entropy loss
By doing so, we are able to establish a robust baseline for ALPL that can be seamlessly integrated into various pool-based AL frameworks. Though encouraging and effective, ALPL with RC loss, similar to all AL frameworks, confronts the inevitable \emph{overfitting} challenge~\cite{chen2006empirical,perez2017effectiveness,shorten2019survey} during the training process with simply few annotated samples provided. Besides, this simple baseline also falls short of the selection of the representative samples with partial labels during the query process.

\begin{figure*}[t]
    \centering
\includegraphics[width=1.0\textwidth]{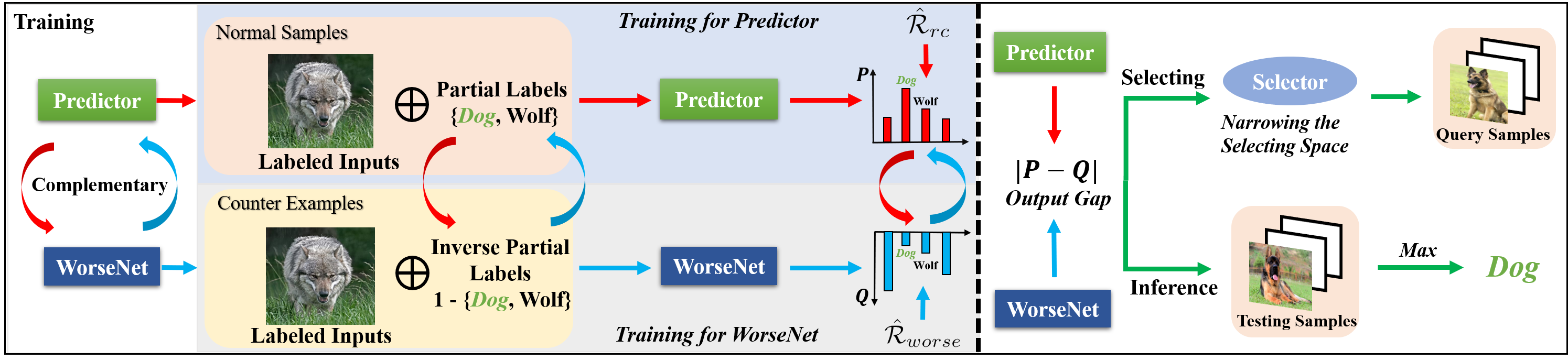}
    \vspace{-6mm}
    \caption{The overall framework of our proposed method to address ALPL. A strong baseline for ALPL is achieved by directly using RC loss to train the predictor (red arrows). To further improve the performance, we propose WorseNet (blue arrows) to extract the useful knowledge from the constructed \emph{counter examples}, individually learning in a complementary way to the predictor. With the help of the distribution gap between the predictor and WorseNet, the selecting and inference process (green arrows) in ALPL could be improved in an explicit way.}
    \label{Fig_framework}
\vspace{-4mm}
\end{figure*}

To move toward better prediction, we turn to an interesting concept from cognitive science named \emph{counter-examples} (CEs). According to the \emph{mental models} in cognitive science~\cite{mental1,mental2,mental3}, humans are able to assess the deductive validity of inference with the help of CEs, leading to drawing an accurate conclusion. 
Inspired by such an adversarial working mechanism, we aim to excavate useful knowledge from CEs to address ALPL by  guiding the predictor to deduce in an explicit way.
% we draw an interesting question that whether CEs could help the predictor in ALPL to improve the recognition accuracy. To select the useful knowledge from CEs,
%relieving the overfitting problem. 
% Based on this motivation, we aim to explore and exploit the useful knowledge from CEs to address ALPL. Specifically,
Firstly, we construct CEs for the predictor  
by directly reversing their partial labels to the inverse version.
%by using the original annotated partial labels. 
% Note that the labels outside the partial label set must belong to the wrong labels. Therefore, 
%we propose to generate CEs for the selected samples  
Building upon the proposed CEs, we propose a simple but effective WorseNet to learn in a way complementary to the predictor. To this end, we propose Worse loss, which contains the 
\emph{inverse RC} (IRC) loss and the \emph{Kullback-Leibler divergence} (KLD) regularization, to guide WorseNet to learn from the inverse partial labels from CEs. Figure~\ref{Fig_framework} illustrates the overall framework. Compared with the predictor, WorseNet would possess lower confidence toward the labels inside the partial label. 

Based on the complementary learning pattern between WorseNet and the predictor, we propose to take advantage of the predicted probability gap between these two networks to separately improve the evaluating and selecting process (shown in Figure~\ref{Fig_framework}). To improve the predicting accuracy, we treat the class with the maximum distribution gap, rather than the maximum predictor score, as the predicted true label during the evaluation. On the other hand, we propose to enhance the sample selector by focusing solely on labels with positive probability gaps, as these labels predominantly cover the true label. This narrows down the range for calculating the uncertainty score, thereby refining the selection process and reducing uncertainty. Consequently, we propose three new selectors in ALPL by adopting this selecting strategy. Experimental results on benchmark-simulated and real-world datasets validate the effectiveness and superiority of our proposed WorseNet in improving both the selector and the predictor in ALPL. Our main contributions are summarized here:

\begin{itemize}
    % \item We for the first time propose a practical setting for \emph{active learning} (AL), i.e., \emph{active learning with partial labels} (ALPL), where the oracle could provide the query samples with partial labels, economically facilitating the annotation process for the experts.
    \item We for the first time propose a practical setting, i.e., \emph{active learning with partial labels} (ALPL), to economically facilitate the annotation process for the experts. In this way, we provide a solid baseline on top of any AL approach to address ALPL. 
    \item We turn to exploring and exploiting the learning pattern from \emph{counter-examples} (CEs), and propose a simple but effective WorseNet to explicitly improve the predictor and the selector in ALPL in a complementary manner. 
    \item Experimental results on four benchmark datasets and five real-world datasets show that our proposed WorseNet achieves promising performance elation over compared baseline methods, achieving state-of-the-art performance in ALPL.
\end{itemize}

\section{Related Work}\label{sup_sec_related_work}
\subsection{Pool-based Active Learning}
According to the different query types between the oracle and the predictor, \emph{active learning} (AL) normally can be divided into membership query synthesis, stream-based query, and pool-based query~\cite{settles1995active}. Pool-based AL, where the selector decides on the annotated samples from a large pool of unlabeled datasets, has drastically appealed to many scholars from academia and industry because of its huge potential value in practical application. With the development of deep learning, pool-based AL has simultaneously experienced the stage from model-driven to data-driven. 

For the prevailing model-driven category, the selector heavily relies on handcrafted features or metrics to query the data. Uncertainty sampling, as the most used metric for the selector, aims to pick out the samples with low confidence from the predictor. Often, such uncertainty could be modeled in three following ways: the posterior probability of a predicted class~\cite{lewis1994heterogenous}, the margin between posterior probabilities of a predicted class and the secondly predicted class~\cite{roth2006margin}, or the entropy~\cite{luo2013latent}. Furthermore, all these uncertainty metrics could be improved, though time-consuming as it is, by using Monte Carlo Dropout and multiple forward passes based on Bayesian inference~\cite{bald,batchbald}. Some methods also modeled the impacts of the selected sample on the current model through Fisher information~\cite{settles2007multiple}, mutual information~\cite{bald,batchbald}, or expected gradient length~\cite{badge}. Specifically, \cite{badge} proposed to select the samples that were disparate and high magnitude in a hallucinated gradient space constructed by using the model parameters of the predictor. Another important metric for the selector is diversity sampling, which aims to select representative and diverse samples for the predictor to better learn from the datasets. To this end, some methods using discrete optimization~\cite{elhamifar2013convex,yang2015multi} focused on sample subset selection while~\cite{nguyen2004active} aimed at mining out the center points of subsets by clustering. Besides, such informative samples could also be highlighted by measuring the expected output changes~\cite{freytag2014selecting}, or the distribution distance between the unlabeled pool and the selected samples~\cite{coreset}.

The methods in the data-driven category describe that the selector, often equipped with deep models, is trained to automatically learn features or metrics. To learn the auto-feature or auto-metric, some methods adopted a generative model-based selector, such as VAE or GAN, to learn to distinguish unlabeled samples from labeled ones~\cite{vaal,tvaal}. Moreover, some methods turned to adopting or designing data augmentation to help the selector better learn the input space~\cite{parvaneh2022active}. \cite{lloss} introduced an auxiliary deep network, predicting the 
``loss'' of the unlabeled samples, to select the samples with large ``loss'' to help the query process.

\subsection{Active Learning with imperfect oracle}
% Recent some works have been positively engaged on active learning in various settings, including graph-based~\cite{graph1,graph2} or class-imbalance cases~\cite{imbalance1,imbalance2}. 

% To further reduce the expensive annotation effort for the oracle
Most works in AL assumed that the oracle would always yield the accurate label, overlooking that the oracle could practically not be infallible in some real-world applications. Therefore, 
a few researchers have investigated AL with an imperfect oracle, where the oracle could provide a wrong (noise) label to the selected sample~\cite{donmez2008proactive,du2010active,yan2016active,chakraborty2020asking}. Early works~\cite{donmez2008proactive} assumed that there were two oracles in the system with one always returning the correct label, while the other returned an incorrect label with a fixed probability. \cite{du2010active} modeled a human-like oracle that would provide noisy labels for the samples with low confidence from the predictor. \cite{yan2016active} studied a case where the oracle could choose to return incorrect labels or abstain from labeling. Some works~\cite{chakraborty2020asking} focused on active learning with multiple noisy oracles and formed the query process as a constrained optimization problem. In this paper, we work towards a new setting for active learning with simply one imperfect oracle involved in the query process, who would annotate the selected samples with partial labels.

\subsection{Partial-Label Learning}
In this part, we concisely give an introduction to the two mainstream strategies for \emph{partial-label learning} (PLL), i.e., the \emph{averaged-based strategy} (ABS) and the \emph{identification-based strategy} (IBS). This method in this paper belongs to the ABS.

ABS treats all candidate labels equally and then averages the model outputs of all candidate labels for evaluation. Some non-parametric methods \cite{hullermeier2006learning,gong2017regularization} focused on predicting the label by using the outputs of its neighbors. Moreover, some approaches \cite{cour2009learning,yao2020deep} concentrated on leveraging the labels outside the candidate set to discriminate the potential true label. Some recent works~\cite{feng2020provably,lv2020progressive,wen2021leveraged} focused on the data generation process and proposed a classifier-consistent method based on a transition matrix. \cite{wen2021leveraged} proposed a family of loss functions, introducing a leverage parameter to consider the trade-off between losses on partial labels and non-partial labels.

IBS focuses on identifying the most possible true label from the candidate label set to eliminate label ambiguity. Early works treated the potential truth label as a latent variable, optimizing the objective function by the maximum likelihood criterion~\cite{liu2014learnability} or the maximum margin criterion~\cite{yu2016maximum}. Later, many researchers engaged in leveraging the representation information of the feature space to generate the score for each candidate label~\cite{wang2019adaptive,wang2022pico,zhang2022exploiting}. \cite{wang2022pico} turned to a contrastive learning framework to eliminate the label disambiguation and reinforce the feature representation learning.
\cite{zhang2022exploiting} proposed to use the class activation map, discriminating the learning pattern of the classifier, to distinguish the potential true label from the candidate label set.

\section{Preliminaries}
\subsection{Symbols and Notations on Pool-based AL}\label{sec_PAL}
Pool-based AL depicts a learning process where the performance gain of the system is achieved through active interaction between the human and the target predictor. Formally, we are given a bunch of training samples $\sX = \{\vx_i\}_{i=1}^n \in \R^{d}$ with a total number of $n$, which is initially split into a small set of labeled samples $\sL = \{\vx_i\}_{i=1}^{l} \in \R^{d}$ and a large pool of unlabeled samples $\sU = \{\vx_i\}_{i=1}^{u} \in \R^{d}$. Note that here $d$ denotes the input dimension, and $\sU \cup \sL = \sX, \sU \cap \sL = \varnothing $. Let $\sY = \{1,2,...,k\} \in \R$ denote the label space with $k$ classes, and $y_i \in \sY$ denote the ground truth for each $\vx_{i}$. A classifier (predictor) $f: \R^{d} \rightarrow \R^{k}$ is then trained by using the original labeled samples $\sL$. Afterwards, a specifically-designed selector $\Psi(\sL,\sU,f)$ evaluates the samples in $\sU$ and selects $\triangle \sU = \{\vx_i\}_{i=1}^{b} \in \sU$ samples to be labeled by an oracle (human expert). Then samples in $\triangle \sU$ with \emph{oracle-annotated true labels} are added to $\sL$, leading to a group of new labeled samples ($\sL = \sL \cup \triangle \sU$), which are further reused to train the classifier $f$.
This cycle of predictor-oracle-based interaction is repeated continuously until a well-performed metric is achieved or the sampling budget is exhausted. The sampling budget aims to restrict the total number of labeled samples for training the classifier, so the overall size of the sampling budget is denoted as $B$ such that $B << u$.

% The success of AL significantly relies on the design of the selector . 

A well-suited selecting metric $\Psi$ could help elate the performance of the model by using as few labeled examples as possible, achieving a win-win situation for the human oracle and the predictor. \emph{Uncertainty} is one of the most prevailing metrics in active learning, arguing that the oracle-annotated samples are able to confound the model most. To mine out those ``uncertain samples'', the selector firstly calculates the uncertainty score for each sample in $\sU$. Typically there are three simple ways to obtain the uncertainty scores by using the model outputs, which are \emph{minimum confidence uncertainty} (MCU), \emph{minimum margin uncertainty} (MMU) and \emph{entropy uncertainty} (EU). These three metrics can be sequentially expressed as follows~\footnote{In Eq. (\ref{eqa_mmu}), $\max^{1}$ ($\max^2$)  means the (second) maximum item.}:
% \begin{equation}\small
% 	 \vx^{*} = \mathop{\arg\max}\limits_{\vx_i \in \sU} \{ 1 - \mathop{\arg\max}\limits_{y_i \in \sY} p(y_i|\vx_i)\} 
% \end{equation}
% \begin{equation}\small
%     		\vx^{*} = \mathop{\arg\min}\limits_{\vx_i \in \sU} \{ \max \limits_{y_i \in \sY}^{1} p(y_i|\vx_i)- \max \limits_{y_i \in \sY}^{2} p(y_i|\vx_i)\} 
% \end{equation}
% \begin{equation}\small
%     		\vx^{*} = \mathop{\arg\max}\limits_{\vx_i \in \sU} \{ \sum \limits_{y_i \in \sY} p(y_i|\vx_i)logp(p(y_i|\vx_i)) \} 
% \end{equation}
\begin{gather}
\small
    	 \vx^{*}_{\text{MCU}} = \mathop{\arg\max}_{\vx_i \in \sU} \{ 1 -  \mathop{\arg\max}_{y_i \in \sY} P(y_i|\vx_i)\}, \label{eqa_mcm} \\
    	     		\vx^{*}_{\text{MMU}} = \mathop{\arg\min}_{\vx_i \in \sU} \{ \max^{1}_{y_i \in \sY}  P(y_i|\vx_i)- \max_{y_i \in \sY}^{2} P(y_i|\vx_i)\}, \label{eqa_mmu}\\
    	   \vx^{*}_{\text{EU}} = \mathop{\arg\max}_{\vx_i \in \sU} \{ \sum \nolimits_{y_i \in \sY} P(y_i|\vx_i)\log(P(y_i|\vx_i))\} , \label{eqa_eu}
\end{gather}
where $P(y_i|\vx_i)$ refers to class-conditional probability and $\vx^{*}$ denotes the selected uncertain samples. Consequently, uncertainty samples handed over to the oracle could be picked by ranking the uncertainty score of each sample in $\sU$ in descending order, resulting in a new labeled dataset to retrain the classifier.

\subsection{Symbols and Notations on PLL}\label{sec_pll}
Formally, let us denote ${\sC} = \{2^{\sY} \backslash \varnothing \backslash {\sY}\}$ as the candidate label space where $2^{\sY}$ is the power set of $\sY$, and $|\sC| = 2^k - 2$ means that the candidate label set is neither the empty set nor the whole label set. For each training instance $\vx_i$,
let ${S_i} \in {\sC}$ be the partial labels. We denote $P(\vx,y)$ and $P(\vx, S)$ as the probability densities of fully labeled examples and partially labeled examples. Building upon the critical assumption of PLL that the candidate label set of each instance must include the correct label, we have ${y_i} \in {S_i}$.
% \begin{equation}\small
% P({y_i} \in {S_i}  | y={y_i}, \vx = {\vx_i} ) = 1,  \forall {y_i} \in {\sY},  \forall {S_i} \in {\sC}.
% \end{equation}
PLL targets at learning a predictor $f$ with training examples sampled from $P(\boldsymbol{x}, S)$ to make correct predictions for test examples. 
% To the best of our knowledge, most methods focus on designing the loss by considering all candidate labels or iteratively extracting the true label from a theoretical perspective. 
Practically, there are two common ways to generate the partial label sets: \textit{(\uppercase\expandafter{\romannumeral1})} \textit{uniformly sampling strategy} (USS). Uniformly sampling the partial label for each training instance from all the possible candidate label sets~\cite{feng2020provably,zhang2022exploiting}. \textit{(\uppercase\expandafter{\romannumeral2})} \textit{Flip Probability Strategy} (FPS). By setting a flip probability $q$ to any false label, the false label could be selected as a candidate label with a probability $q$ \cite{feng2019partial,yan2020partial,lv2020progressive,wen2021leveraged,wang2022pico}. In this paper, we adopt both of them to generate partial labels. Refer to the \textbf{Appendix} file for more details.

\section{Active Learning with partial labels}\label{sec_new_setting}
In this section, we introduce in detail a new but practical setting based on AL, namely \emph{active learning with partial labels} (ALPL). Different from the previous AL settings, which may be impractical and demanding for the oracle, requiring the oracle to provide the true labels~\cite{bald,tran2019bayesian,tvaal} to the selected samples, ALPL regulates that the oracle is asked to label the samples with partial labels that are widely used in real-world scenarios~\cite{cour2011learning,zeng2013learning}. Compared with AL, ALPL eases the annotation pressure for the oracle when facing confusing samples. Therefore, we believe that ALPL is full of research significance, and  a formal definition of ALPL is given as

\noindent\textbf{Definition of ALPL.} \emph{Active learning with partial labels (ALPL) trains a predictor with initial training samples annotated with partial labels, uses its selector to select the samples from the unlabeled samples, sends them to an oracle who only provides partial labels, adds them into the labeled training samples, and then retrains the predictor. }

Figure~\ref{Fig_motivation} illustrates the pipelines of AL and our proposed ALPL. Note that the key difference between ALPL and AL is the label supervision, so it is intuitive to address ALPL by simply adopting a PLL-based loss function to train the predictor, relieving the negative effects caused by the false positive labels in the candidate label sets. In this case, we use RC loss~\cite{lv2020progressive,feng2020provably}, as one of the most prevailing state-of-the-art loss functions~\cite{wen2021leveraged,wang2022pico,zhang2022exploiting}, to address ALPL in a simple but effective manner. 
% RC loss could guide the predictor to approach the true label by re-weighting the importance of each label in candidate label sets, where 
The empirical risk function $\mathcal{\hat R}_{\mathrm{rc}}$ is defined as
\begin{equation}\small\label{eqa_rcloss}
     \mathcal{\hat R}_{\mathrm{rc}}  = \sum \nolimits_{i = 0}^{l}\sum \nolimits_{j \in S_i} \frac {P(y_i = j|\vx_i)}  {\sum \nolimits_{z \in S_i} P(y_i = z|\vx_i) } \mathcal{L}(f(\vx_i),j).
\end{equation}
Here $\mathcal{L}(f(\vx),s), s \in S$ refers to the cross entropy loss. As shown in Eq.~(\ref{eqa_rcloss}), RC loss is essentially a form of weighted cross entropy among the labels in the candidate set, which is theoretically proved to reach risk consistency in PLL, i.e., achieving comparable performance when compared to the fully supervised methods. Therefore, here we train the predictor $f$ with RC loss to serve as the baseline of ALPL. In this way, we could seamlessly apply any AL-based frameworks to address ALPL (ten approaches implemented in our paper, see Section~\ref{sec_experiments} for more details).

\section{WorseNet: learning from Counter Examples}
In this section, we introduce our proposed method to address ALPL in detail. Figure~\ref{Fig_framework} illustrates the overall framework of our proposed WorseNet. Section~\ref{sub_section_Constructing counter-examples} introduces the training procedure of our WorseNet. Section~\ref{sub_section_Predicting better with WorseNet} and Section~\ref{sub_sec_Selecting better with WorseNet} introduce how WorseNet could address ALPL in both prediction and selection processes.

\subsection{Constructing Counter-Examples}\label{sub_section_Constructing counter-examples}
Though effective, it is observed two potential issues for the baseline method in ALPL. The first goes to the \emph{overfitting}~\cite{chen2006empirical,perez2017effectiveness,shorten2019survey}, which is a common challenge in both AL and ALPL due to the utilization of a relatively small set of annotated samples. % Due to simply few samples used in the training round, 
Meanwhile, the sample selection process, as the fundamental part of ALPL, aims to select representative samples that are successively annotated with partial labels, and such distinction sets ALPL apart from conventional AL. 

% As adopted in most works,  could typically be a simple but effective choice to relieve such an overfitting problem. Nevertheless, we contend that it would be difficult to improve the classifier by either introducing one or more DA operations to all training samples, or designing customized DA operation to each single training sample. 
To address these two problems, we turn to an interesting concept in human reasoning. When humans perceive and learn the world, vision yields a mental model to help understand the things described in the scene, and builds a prior knowledge base to proceed further reasoning. Specifically, when evaluating the deductive validity of an \emph{inference}, humans search for \emph{counter-examples} (CEs) to help disapprove the conjecture~\cite{mental1,mental2,mental3}. For instance, the fact that ``John Smith is not a lazy student'' is one CE to the \emph{inference} ``all students are lazy''. Therefore, we can tell that ``all students are lazy'' is a false conclusion because of ``John Smith''. Intuitively, CEs occupy on an important position in human reasoning. Inspired by the effectiveness of CEs in the mental model, we are driven to draw an interesting question: \emph{can the predictor also benefit from CEs}? Thus, here we aim to explore and exploit CEs from the data, explicitly assisting the predictor to improve its performance in ALPL.

The first question goes to how to construct CEs for the predictor. It is emphasized that CEs rigorously deplore the \emph{inference}. Let us consider that we classify an image of a dog with a one-hot label, and assume that the \emph{inference} here is ``The image has a dog''. In this way, this conjecture is rejected once this image is annotated ``0'' at the ``Dog'' index. Here the simple inverse on the true label intuitively leads to a CE, which violates the original accurate \emph{inference}, leading to a complementary conclusion. Motivated by this, we propose to build up CEs for the predictor by adopting label inversion to the selected samples. Formally, we are given a set of data samples $\sW = \{\vx_i\}_{i=1}^l \in \R^{d}$  such that $\sW = \sL$, and the assigned label of each sample in $\sW$ is defined as follows:
\begin{equation}\small\label{eqa_wrong_label}
\overline{S}_i = \sY - S_i,    
\end{equation}
where $\overline{S}_i$ denotes the candidate label set for the instance in $\sW$. Intuitively, $\overline{S}_i$ is complementary to $S_i$, i.e., $\overline{S}_i = \complement_\sY S_i$, meaning that there is no true label within $\sW$. For convenience, we name the candidate label set $\overline{S}$ as the \emph{inverse partial label} (IPL). Note that IPL is different from the  \emph{complementary label}~\cite{ishida2017learning}. The former provides a wrong indicator to the samples while the latter aims to train a true-label predictor by specifying the classes
that the example does not belong to.  

There are two benefits to forming IPL by following Eq. (\ref{eqa_wrong_label}) in ALPL. Firstly, it is convenient and efficient to construct CEs with a label-based operation to the selected label samples $\sL$. Secondly, IPL considers that all false labels outsides $\overline{S}_i$ shall become the inverse knowledge to the instance $\vx_i$, enriching the label variety of CEs. 
% Overall, our proposed IPL takes account of both operational simplicity and label diversity to make CEs in ALPL. 

\subsection{Predicting better with WorseNet}\label{sub_section_Predicting better with WorseNet}
In this section, we introduce how to assist the predictor with the help of the proposed CEs in ALPL. Firstly, an extra classifier apart from the predictor is needed to learn from CEs obtained from $\sW$ annotated with IPL. Formally, let us name such a classifier as the WorseNet and denote it as $w: \R^{d} \rightarrow \R^{k}$. Note that $w$ shares the same input and output space as the predictor $f$ since $w$ is trained with training samples from $Q(\vx, \overline{S})$, which denotes the probability densities of samples with IPL. To help $w$ extract the inverse knowledge from $Q(\vx, \overline{S})$, we formulate this learning process, treating the IPL as the normal partial labels, to a similar PLL problem, where we propose \emph{inverse RC} (IRC) loss to address it as follows: 
\begin{equation}\small\label{eqa_wrcloss}
     \mathcal{\hat R}_{\mathrm{irc}} =  \sum \nolimits_{i=1}^{l} \sum \nolimits_{j \in \overline{S}_i} \frac {Q(y_i = j|\vx_i)}  {\sum _{z \in \overline{S}_i} Q(y_i = z|\vx_i) } \mathcal{L}(w(\vx_i),j),
\end{equation}
where $\mathcal{\hat R}_{\mathrm{irc}} (\mathcal{L}, w)$ denotes the empirical risk function for $w$, and $Q(y|\vx)$ denotes the class-conditional probability modeled by $w$. Clearly, IRC loss focuses on the labels outside the candidate label set in a way complementary to RC loss. 

Supported by the IRC loss, WorseNet is able to latch on to a pattern that is complementary to the predictor. To improve the predictor with WorseNet, we leverage the output distribution gap between $w$ and $f$ to predict the true label during the inference. Since the original true label only lies in the candidate label set $S$, we should intuitively aim at enlarging the gap of the output distribution on $S$ between $f$ and $w$. To this end, we further add a \emph{Kullback-Leibler divergence} (KLD) regularization item for $w$, regulating its learning process toward the gainful direction to the predictor. Specifically, the KLD item is expressed as
\begin{equation}\small\label{eqa_kld}
     \mathrm{KLD} =   \sum \nolimits_{i=1}^{l} \sum \nolimits_{j \in \overline{S}_i} {P(y_i = j|\vx_i)} \log \frac {{P(y_i = j|\vx_i)}} {{Q(y_i = j|\vx_i)}}.
\end{equation}

Note that here we stop the gradient backpropagation of $P$ when training $w$. As shown in Eq.~(\ref{eqa_kld}), we calculate the KLD between the predictor and WorseNet by merely using their outputs inside $\overline{S}$, which could be minimized to implicitly enlarge the output distribution of the candidate set between $f$ and $w$. In all, the learning loss function for WorseNet, denoted as Worse loss, could be expressed as follows:
\begin{equation}\small\label{eqa_wloss}
     \mathcal{\hat R}_{\mathrm{worse}} = \mathcal{\hat R}_{\mathrm{irc}} (\mathcal{L}, w) + \alpha\mathrm{KLD}.
\end{equation}
where $\alpha$ is a regularized parameter and we empirically set $\alpha=1$. After training by Eq. (\ref{eqa_wloss}), the predictor during the inference could predict the potential true label by
\begin{equation}\small\label{eqa_wlabel}
     y_i^{*} = \mathop{\arg\max}\nolimits_{y_i \in \sY} \{ P(y_i|\vx_i) + (1- Q(y_i|\vx_i))\},
\end{equation}

where $y_i^{*}$ denotes the predicted true label of $\vx_i$. Note that here we use $1-Q$ to help the predictor recognize the true label. As WorseNet is trained independently of the predictor, the proposed WorseNet is able to benefit the predictor on top of any selector in ALPL. To better illustrate this, we provide the following theorem. 

\begin{theorem}\label{theorem1}
Assume that the posterior probability of WorseNet satisfies $Q(y = j | \vx_i) + P(y = j | \vx_i) = 1$ for any label $j \in {\sY}$ of sample $\vx_i$, and the loss function $\mathcal{L}$ is the standard cross entropy. Then the Worse loss $\mathcal{\hat R}_{\mathrm{worse}}$ holds
\begin{equation}\small\label{eqa_theorem}
    \mathcal{\hat R}_{\mathrm{worse}} \propto 
    {\sum\nolimits_{i=1}^{l}} \sum\nolimits_{j \in \overline{S}_i} -n(Q_{ij}) \mathrm{log}(Q_{ij}),   
\end{equation}
\end{theorem}
% \noindent\textbf{Theorem 1.} \label{ref_theorem}
% \emph{Assume that the posterior probability of WorseNet satisfies $Q(y = j | \vx_i) + P(y = j | \vx_i) = 1$ for any label $j \in {\sY}$ of sample $\vx_i$, and the loss function $\mathcal{L}$ is the standard cross entropy. Then the Worse loss $\mathcal{\hat R}_{\mathrm{worse}}$ holds}
% \begin{equation}\small\label{eqa_theorem}
%     \mathcal{\hat R}_{\mathrm{worse}} \propto 
%     {\sum\nolimits_{i=1}^{l}} \sum\nolimits_{j \in \overline{S}_i} -n(Q_{ij}) \mathrm{log}(Q_{ij}),   
% \end{equation}
where $Q_{ij}$ represents $Q(y = j | \vx_i)$ for simplicity and $n(Q_{ij}) > 0, \forall {Q_{ij}} \in [0,1]$. The proof and analysis of Theorem \ref{theorem1} is in the \textbf{Appendix} file. Theorem \ref{theorem1} shows that the WorseNet is learned to approximate the false labels in $\overline{S}$ in an entropy-based manner. As $\mathcal{\hat R}_{\mathrm{worse}}$ decreases and $Q_{ij} \rightarrow 1$, the predictor is correspondingly pushed away from $\mathcal{\overline{S}}$ ($P_{ij} \rightarrow 0$). In all, the Worse loss could serve as an auxiliary module to the predictor by considering the extra supervision on the elements outside the partial labels. For convenience, we denote this improvement of WorseNet to the predictor during the evaluation as WorseNet-Predictor (WP), and its pseudo-code is given in Algorithm \ref{alogro_WorseNet}. 

\subsection{Selecting better with WorseNet}\label{sub_sec_Selecting better with WorseNet}
% Here we would like to emphasize that the essence of AL or ALPL is the sampling strategy of the selector, properly choosing the labeled data samples to help the predictor to gain as much as possible. Consequently, 
In this section, we illustrate that the proposed WorseNet can also promote the sampling metric of some uncertainty-based selectors. As shown in Section~\ref{sec_PAL}, a selector $\Psi(\sL,\sU,f)$ needs to calculate the uncertainty score of $\vx_i$ in the entire class space since it has no prior knowledge about the class of this sample. We argue that such a strategy could be further improved if the class space for obtaining the uncertainty could be narrowed down, bringing well inductive bias to the selector. 
\begin{figure}[!t]
\vspace{-5mm}
\centering
% \begin{minipage}{1.0\linewidth}
\begin{algorithm}[H]
\small
	\caption{ALPL with WorseNet-Predictor (WP)}
	\label{alogro_WorseNet}
	\renewcommand{\algorithmicrequire}{\textbf{Input:}}
	\renewcommand{\algorithmicensure}{\textbf{Output:}}
	\begin{algorithmic}[1]
		\REQUIRE Predictor $f$, WorseNet $w$, iterations $T$, unlabeled examples $\sX$, an oracle $\mathcal{O}$, a selector  $\Psi(\sL,\sU,f)$, initial sampling size $b_0$, query size $b$, sampling budget $B$.
        \STATE  {\textbf{Label} $b_0$ samples drawn uniformly at random from $\sX$ with partial labels $S$, forming the initial labeled samples $\sL$, and all the remaining samples in $\sX$ compose the unlabeled samples $\sU$;}
        
        \STATE  {\textbf{Train} an initial $f$ on $\sL$ by $\mathcal{\hat R}_{\mathrm{rc}}$ in Eq. (\ref{eqa_rcloss}); }
        \STATE  {\textbf{Label} the samples from $\sL$ with IPL $\overline{S}$ by Eq. (\ref{eqa_wrong_label}), forming the initial CEs $\sW$;}
        
        \STATE  {\textbf{Train} an initial $w$ on $\sW$ by $\mathcal{\hat R}_{\mathrm{worse}}$ in Eq. (\ref{eqa_wloss});}

        \WHILE {$t < T$ and $B > 0$} 
        \STATE {\textbf{Select} $b$ samples from $\sU$ by $\Psi(\sL,\sU,f)$, building the query samples $\triangle \sU$;}
        \STATE {\textbf{Label} $\triangle \sU$ with $S$ by $\mathcal{O}$, forming the labeled query samples $\triangle \sL$; }
        \STATE {\textbf{Label} $\triangle \sU$ with $\overline{S}$ by Eq. (\ref{eqa_wrong_label}), forming the IPL-annoatated query samples $\triangle \sW$; }
        \STATE { $ \sU\Leftarrow \sU - \triangle \sU$; $\sL \Leftarrow \sL \cup \triangle \sL$; $\sW \Leftarrow \sW \cup \triangle \sW$;}
        \STATE  {\textbf{Train} $f$ on $\sL$ labeled with $S$ by $\mathcal{\hat R}_{\mathrm{rc}}$ in Eq. (\ref{eqa_rcloss});}
        \STATE  {\textbf{Train} $w$ on $\sW$ labeled with $\overline{S}$ by $\mathcal{\hat R}_{\mathrm{worse}}$ in Eq.
        (\ref{eqa_wloss});}
    \STATE { $ t \Leftarrow t +1$; $B \Leftarrow B - b$;}
    \ENDWHILE
                \STATE  {\textbf{(Inference):} Predict the true label $y^{*}$ in Eq. (\ref{eqa_wlabel}). }
        \ENSURE $f,w$.
	\end{algorithmic}
	
\end{algorithm}
\vspace{-8mm}
% \end{minipage}
\end{figure}
As shown in Eq.~(\ref{eqa_wlabel}), we test our proposed framework during the inference by measuring the gap of the output distribution between $f$ and $w$. In particular, we assume that the true label is the class with the maximum probability distance between $f$ and $w$. As $f$ focuses on the candidate label set $S$ while $w$ learns from CEs, the former one shall have a higher response to the labels in $S$ than the latter one. Hence, it reveals that the potential true label must satisfy $P > Q$ since the true label absolutely lies on $S$. Based on this, we construct a pseudo partial label candidate set $S^{'}$ for each unlabeled sample in $\sU$ as follows:
\begin{equation}\small\label{eqa_newS}
     S_i^{'} = \{z | P(y_i = z|\vx_i) - Q(y_i = z | \vx_i) \geq     0, z \in \sY \}.
\end{equation}
Building upon $S^{'}$, a selector could narrow the class range of acquiring the uncertainty score in $\sU$. To this end, we propose three sampling strategies based on MCU (Eq.~(\ref{eqa_mcm})), MMU (Eq.~(\ref{eqa_mmu})), and EU (Eq.~(\ref{eqa_eu})) by directly substituting $\sY$ with $S{'}$. For convenience, we denote the improvement of WorseNet on the selector as WorseNet-Selector (WS), and denote these three methods as WS-MCU, WS-MMU, and WS-EU.

\section{Experiments}\label{sec_experiments}
In this section, we evaluate our proposed WP, WS-MCU, WS-MMU, and WS-EU against several algorithms from the literature, and extensive experiments are implemented to verify the correctness and effectiveness of our proposed modules. More details could be found in the \textbf{Appendix} file.

\subsection{Benchmark datasets comparisons}\label{sec_benchmark_datasets_comparisons}
\noindent\textbf{Datasets and backbones.} Our proposed WorseNet-based modules are evaluated on four popular benchmark datasets, which are MNIST~\cite{minist}, Fashion-MNIST~\cite{fashion}, SVHN~\cite{svhn} and CIFAR-10~\cite{cifar}. Note that it is necessary for the oracle to manually generate the candidate label sets for these datasets, which are supposed to be used for single-classification problems. Recall that we introduce two different candidate label generation approaches, i.e., USS and FPS. For FPS, we set $q \in \{0.3,0.5\}$ to represent different ambiguity degrees. For MNIST and Fashion-MNIST, we adopt a 3-layer MLP and a simple CNN-based network denoted as C-Net (similar to the network used in \cite{bald,batchbald}) as the backbones for the predictor. For SVHN and CIFAR-10, we follow most works~\cite{lloss,badge,tvaal} and choose ResNet18~\cite{resnet} and VGG11~\cite{vgg} as the base models. Note that WorseNet $w$ follows the identical architecture to the predictor $f$.

\noindent\textbf{Compared methods and training settings.} We compare our proposed modules with ten approaches which contains seven model-driven methods: 1) Random Sampling (RS), 2) MCU, 3) MMU, 4) EU, 5) Coreset~\cite{coreset}, 6) BALD~\cite{batchbald}, 7) BADGE~\cite{badge}, and three data-driven methods: 8) LL4AL~\cite{lloss}, 9) VAAL~\cite{vaal} and 10) TA-VAAL~\cite{tvaal}. For the seven model-driven methods, we adopt the Adam optimizer~\cite{adam} with a learning rate of $0.001$ to train $f$. We take a mini-batch size of $256$ images and train all seven methods for $200$ epochs. For three data-driven methods, we strictly follow the reported training hyper-parameters in their papers \cite{lloss,vaal, tvaal}. Besides, we simply adopt ResNet18 as the backbone for $f$ and $w$ in these three data-driven methods. For the ALPL setting, we construct an initial labeled set $\sL$ with the size $b_0 = 20$, and acquire $b = 100$ instances ($b = 1000$ for SVHN and CIFAR-10) from $\sU$ in each query round, following prior works~\cite{bald,batchbald,lada}. We repeat the query process $10$ times such that the overall budget size $B = 1000$ ($B = 10000$ for SVHN and CIFAR-10). Note that we directly adopt RC loss on these ten methods to build the baselines (see Section~\ref{sec_new_setting} for more details). To guarantee comparison fairness, we repeatedly conduct all experiments 5 times and report the average test accuracy using the model achieving the maximum performance on a validation set, which is constructed by randomly selecting $100$ instances from the training datasets. Here the validation performance of $w$ is measured by Eq.~(\ref{eqa_wlabel}). All the implemented methods are trained on 2 RTX3090 GPUs each with 24 GB memory.

\noindent\textbf{Experiment results.} As shown in Table \ref{tab_bench_FPS_0.5}, following the default settings, our proposed WorseNet shows its effectiveness and superiority in addressing ALPL on these four benchmark datasets. Firstly, WP can bring a constant gain to the classifier regardless of the backbone and the adopted AL methods. Moreover, the improvement by WP shall be witnessed in both USS and FPS cases, validating that our WP does not rely on any data generation assumption. Our approach could also deliver promising performance with full access to the datasets, which means that WP is also an effective way to address PLL. Particularly, we would like to highlight 
a counter-intuitive phenomenon that RS may perform better than some methods in some cases. RS (70.73\%) performs far better than EU (64.58\%) and Coreset (53.17\%) in Fashion-MNIST. This counter-intuitive could also be seen in~\cite{lada,lloss,vaal,badge}. This phenomenon can be attributed to the instability caused by a relatively small number of labeled samples.

% \begin{table*}[!htbp]
% \centering
% \caption{Testing performance with different data augmentation. The \underline{underline} points out the improved accuracy by WP. Other settings are similar with Table~\ref{tab_bench_FPS_0.5}.}
% \vspace{-4mm}
% \label{tab_dataaug}
% \resizebox{0.48\textwidth}{!}{
% % \setlength{\tabcolsep}{4mm}{
% \begin{tabular}{c|c}
% \toprule
% Data Augmentation + RS & Fashion-MNIST \\
% \midrule
% Random Erasing & 66.15 $\pm$ 5.98 / \underline{67.12 $\pm$ 4.25}  \\ 
% Mixup & 16.15 $\pm$ 4.67 / \underline{16.56 $\pm$ 4.87} \\ 
% Random Crop & 71.29 $\pm$ 1.00 / \underline{72.82 $\pm$ 0.51}
% \\
% \bottomrule
% \end{tabular}
% % }
% }
% \vspace{-3.5mm}
% \end{table*}

\begin{table*}[h]
\centering
\caption{Test performance of the methods on benchmark datasets using label generation by FPS ($q = 0.5$). The best results are marked in \textbf{bold}. -/+ WP denotes whether the predictor is helped by WorseNet. The \underline{underline} points out improved accuracy by WP. \textcolor[RGB]{0,204,102}{$\uparrow$} indicates the improved accuracy is beyond 1\%. The backbones for MNIST and Fashion-MNIST are C-Net, and for SVHN and CIFAR-10 are ResNet18. \emph{Fully-supervised PLL} denotes the training performance with full 
access to the partially labeled datasets. Here the standard deviation is ignored. }
\label{tab_bench_FPS_0.5}
\resizebox{1.0\textwidth}{!}{
\setlength{\tabcolsep}{4mm}{
\begin{tabular}{c|c|c|c|c}
\toprule
Methods ( -/+ WP) & MNIST & Fashion-MNIST & SVHN & CIFAR-10 \\
\midrule
{RS} & 89.26 / \underline{90.11}
&70.73 / \underline{71.17} 

& 71.63 / \underline{72.23} 
& 54.57 / \underline{55.41}  \\   

{MMU} & \hspace{1.8mm}  95.18 / \underline{96.37} 

\textcolor[RGB]{0,204,102}{$\uparrow$}
& \hspace{1.8mm}   74.22 / \underline{76.44}  

\textcolor[RGB]{0,204,102}{$\uparrow$}
& \hspace{1.8mm}  75.13 / \underline{76.21}  

\textcolor[RGB]{0,204,102}{$\uparrow$}
& \hspace{1.8mm}  57.65 / \underline{58.67} 

\textcolor[RGB]{0,204,102}{$\uparrow$}
\\    

{MCU} & 93.75 / \underline{94.68}  
& \hspace{1.8mm}  64.59 / \underline{65.75}  

\textcolor[RGB]{0,204,102}{$\uparrow$}
&76.28 / \underline{77.09} 

& \hspace{1.8mm} 58.41 / \underline{59.51} 

\textcolor[RGB]{0,204,102}{$\uparrow$}
\\

{EU} & 90.83 / \underline{91.28} 
&  64.58 / \underline{65.16} 
&  75.17 / \underline{76.08} 
&  \hspace{1.8mm} 57.58 / \underline{58.79}  
\textcolor[RGB]{0,204,102}{$\uparrow$}
\\    

{Coreset} & \hspace{1.8mm}  86.05 / \underline{87.65} 

\textcolor[RGB]{0,204,102}{$\uparrow$}
&  \hspace{1.8mm}  53.14  / \underline{61.62} 

\textcolor[RGB]{0,204,102}{$\uparrow$}
&  \hspace{1.8mm} 75.32 / \underline{76.10} 
\textcolor[RGB]{0,204,102}{$\uparrow$}

&  \hspace{1.8mm} 59.25 / \underline{60.37} 

\textcolor[RGB]{0,204,102}{$\uparrow$}
\\    

{BALD} &  \hspace{1.8mm} 94.08 / \underline{95.11} 

\textcolor[RGB]{0,204,102}{$\uparrow$}
&  \hspace{1.8mm}  70.95 / \underline{72.95}  

\textcolor[RGB]{0,204,102}{$\uparrow$}
&   77.15 / \underline{77.82} 

& \hspace{1.8mm}  59.09 / \underline{60.13}    

\textcolor[RGB]{0,204,102}{$\uparrow$}
\\  

{BADGE} & 96.01 / \underline{96.49} 
& 76.75 / \underline{77.10}  
&  \hspace{1.8mm}  77.23 / \underline{78.76}
\textcolor[RGB]{0,204,102}{$\uparrow$}
& \hspace{1.8mm} 59.04 / \underline{60.30}   
\textcolor[RGB]{0,204,102}{$\uparrow$}
\\  

{LL4AL} & 81.91 / \underline{82.75}
& 60.91 / \underline{61.62} 
& \hspace{1.8mm} 76.69 / \underline{77.80}  
\textcolor[RGB]{0,204,102}{$\uparrow$}

&  \hspace{1.8mm} 55.81 / \underline{56.97} 

\textcolor[RGB]{0,204,102}{$\uparrow$}
\\ 

{VAAL} & 90.68 / \underline{91.08}  
& 75.18 / \underline{75.44} 
& 77.81 / \underline{78.05}  
& 56.69 / \underline{57.32}  
\\ 

{TA-VAAL} & 90.93 / \underline{91.26} 
& 75.21 / \underline{75.90} 
& 78.07 / \underline{78.40}  
& \hspace{1.8mm}  56.81 / \underline{57.94}  

\textcolor[RGB]{0,204,102}{$\uparrow$}
\\ 
\midrule
{WS-MMU} & 95.74 / \underline{\textbf{96.66}} 
& 77.08 / \underline{\textbf{77.75}} 
& 77.51 / \underline{78.18} 
&  58.63 / \underline{59.36}   

\\ 
{WS-MCU} & 94.96 / \underline{95.17} 
& \hspace{1.8mm}  68.36 / \underline{69.77} 

\textcolor[RGB]{0,204,102}{$\uparrow$}
& 78.81 / \textbf{\underline{79.61}} 
&\hspace{1.8mm} 59.39 / \underline{60.83}   
\textcolor[RGB]{0,204,102}{$\uparrow$}
\\ 

{WS-EU} & 93.90 / \underline{94.80} 
& \hspace{1.8mm}  66.01 / \underline{67.75} 

\textcolor[RGB]{0,204,102}{$\uparrow$}
& 76.09 / \underline{77.12}
& 58.45 / \underline{\textbf{59.12}}    \\ 

\bottomrule
{\emph{Fully-supervised PLL}} & 97.61 / \underline{98.27} 
& \hspace{1.8mm}  84.49 / \underline{85.87} 

\textcolor[RGB]{0,204,102}{$\uparrow$}
& 92.36 / \underline{93.01}
&\hspace{1.8mm} 71.89 / \underline{73.58}  \textcolor[RGB]{0,204,102}{$\uparrow$}  \\ 
\bottomrule
\end{tabular}
}
}
\end{table*}

\begin{figure*}[!htbp]
\vspace{-4mm}
\centering
\includegraphics[width=0.8\linewidth]{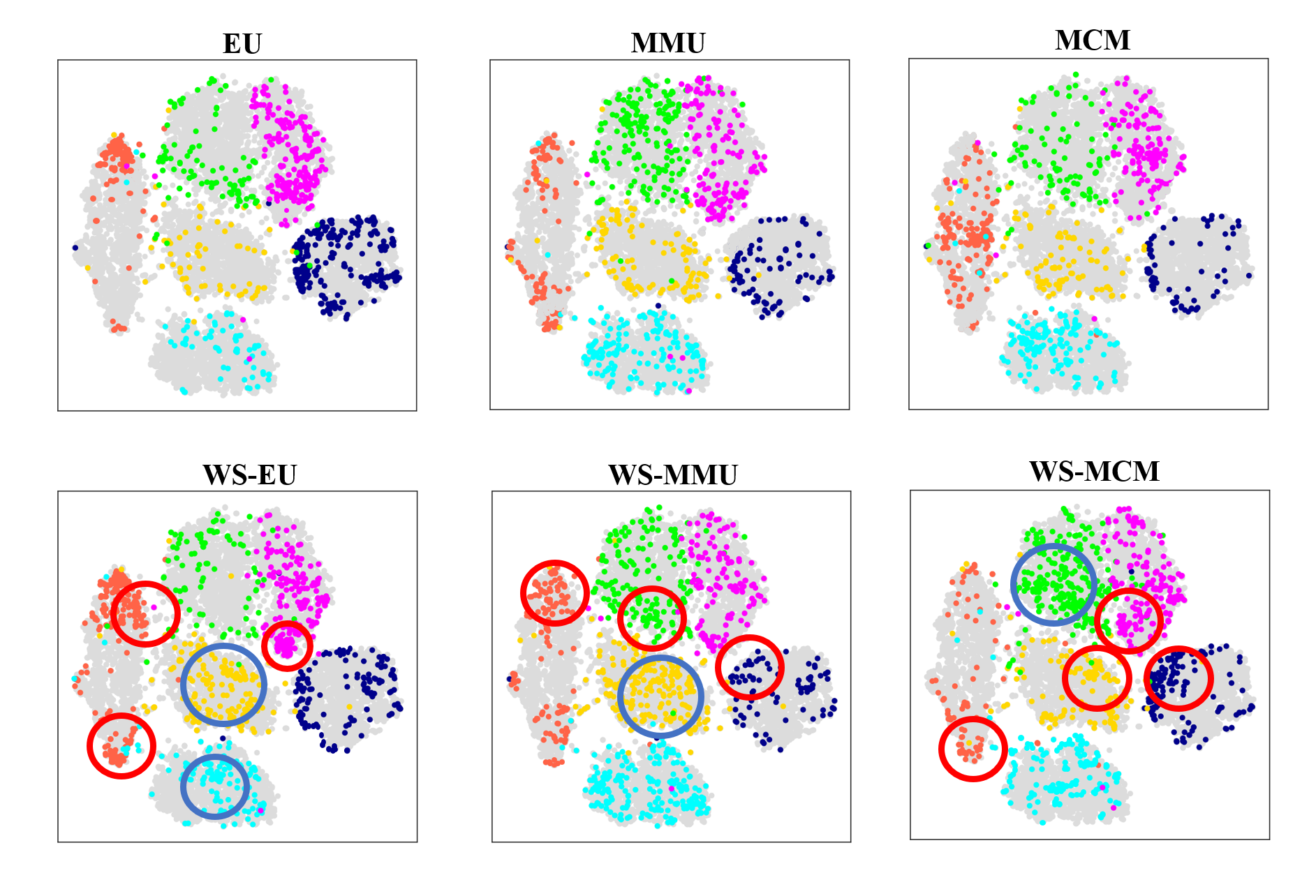}
    \caption{Visualized tSNE results of EU and WS-EU in MNIST with FPS ($q=0.5$). The red circles mark that more samples near the class boundary are selected, and the blue circle mark that more samples near the center of the class cluster are selected. }
    \label{Fig_tsne}
\vspace{-3mm}
\end{figure*}

For three WS-based selectors, i.e., WS-MMU, WS-MCU, and WS-EU, they are found to better elate the performance of the classifier in ALPL when compared to the original version. Additionally, these three improved uncertainty-based approaches show competitive performance compared with the other ten AL methods, and such performance could be further improved by reusing WP to reach state-of-the-art performance in ALPL. As shown in Figure~\ref{Fig_tsne}, we select 6 classes and visualize the selected samples of EU and WS-EU. Compared to EU, our WS module could enforce the selector to select more representative and diverse samples. Specifically, our proposed selectors are able to select more samples (marked by the red circle) that nearby the class boundary. Besides, more samples near the center of the class cluster are also selected to ensure the accuracy (marked by the blue circle), illustrating that our WS could help ALPL to select more representative samples with partial labels. Overall, the experimental results on four benchmark datasets reasonably verify the generalization
and effectiveness in addressing ALPL. 

\begin{table*}[!t]
\centering
\caption{Test performance of compared methods on five real-world datasets. The \underline{underline} points out improved accuracy by WP. \textcolor[RGB]{0,204,102}{$\uparrow$} indicates the improved accuracy is beyond 3\%. Note that three data-driven methods are not implemented here due to the framework incompatibility.}
\label{tab_RW}
\resizebox{1.0\textwidth}{!}{
\begin{tabular}{c|c|c|c|c|c}
\toprule
Methods ( -/+ WP) & Lost & MSRCV2 & BirdSong & SoccerPlayer & Yahoo!News \\
\midrule
{RS} & \hspace{1.8mm}  51.68 / \underline{55.93} \textcolor[RGB]{0,204,102}{$\uparrow$}
& \hspace{1.8mm}  40.91 / \underline{44.59} \textcolor[RGB]{0,204,102}{$\uparrow$}
& \hspace{1.8mm}  57.01 / \underline{62.05} 

\textcolor[RGB]{0,204,102}{$\uparrow$}
& 48.31 / \underline{49.94} 
& \hspace{1.8mm}  52.59 / \underline{56.20}

\textcolor[RGB]{0,204,102}{$\uparrow$}
\\

{MMU} & \hspace{1.8mm} 53.58 / \underline{56.75}  
\textcolor[RGB]{0,204,102}{$\uparrow$}

&  44.32 / \underline{46.59} 
& \hspace{1.8mm}  58.60 / \underline{62.80} 

\textcolor[RGB]{0,204,102}{$\uparrow$}
& 50.17 / \underline{51.92} 

& 56.85 / \underline{59.37}

\\    

{MCU} & 53.06 / \underline{56.25}  
& \hspace{1.8mm}   
43.18 / \underline{46.02}   

\textcolor[RGB]{0,204,102}{$\uparrow$}
& 63.39 / \underline{66.43} 
& 51.32 / \underline{53.06} 

& \hspace{1.8mm}  55.42 / \underline{58.55} 

\textcolor[RGB]{0,204,102}{$\uparrow$}
\\    

{EU} & \hspace{1.8mm}    48.21 / \underline{54.46}
\textcolor[RGB]{0,204,102}{$\uparrow$}

& \hspace{1.8mm} 41.32 / \underline{45.14} 
\textcolor[RGB]{0,204,102}{$\uparrow$}

&  \hspace{1.8mm} 63.22 / \underline{66.60}

\textcolor[RGB]{0,204,102}{$\uparrow$}
&   49.19 / \underline{51.06}

&\hspace{1.8mm} 54.94 / \underline{57.98}
\textcolor[RGB]{0,204,102}{$\uparrow$}
\\    

{Coreset} & \hspace{1.8mm} 52.32 / \underline{56.79}  
\textcolor[RGB]{0,204,102}{$\uparrow$}
&   41.92 / \underline{44.32} 

&  \hspace{1.8mm} 60.15 / 66.43  

\textcolor[RGB]{0,204,102}{$\uparrow$}
& 50.03 / \underline{50.83} 
& 50.98 / \underline{52.02} 
\\    

{BALD} &  52.79 / \underline{54.57}

& \hspace{1.8mm} 40.91 / \underline{47.73} 
\textcolor[RGB]{0,204,102}{$\uparrow$}

&  \hspace{1.8mm} 62.80 / \underline{65.20} 

\textcolor[RGB]{0,204,102}{$\uparrow$}
& \hspace{1.8mm} 48.94 / \underline{52.38}
\textcolor[RGB]{0,204,102}{$\uparrow$}
&  \hspace{1.8mm} 54.21 / \underline{58.24} 

\textcolor[RGB]{0,204,102}{$\uparrow$}
\\  

{BADGE} &  \hspace{1.8mm} 52.00 / \underline{53.79}  

\textcolor[RGB]{0,204,102}{$\uparrow$}
&  \hspace{1.8mm} 50.57 / \underline{\textbf{53.98}}   

\textcolor[RGB]{0,204,102}{$\uparrow$}
&  \hspace{1.8mm} 64.61 / \underline{68.05} 

\textcolor[RGB]{0,204,102}{$\uparrow$}
& 50.72 / \underline{53.47} 
&  \hspace{1.8mm} 57.72 / \underline{\textbf{60.98}} 

\textcolor[RGB]{0,204,102}{$\uparrow$}
\\  

\midrule
{WS-MMU} & \hspace{1.8mm} 54.09 / \underline{\textbf{57.14}} 
\textcolor[RGB]{0,204,102}{$\uparrow$}
& \hspace{1.8mm} 46.59 / \underline{50.00} 

\textcolor[RGB]{0,204,102}{$\uparrow$}
&  \hspace{1.8mm} 62.42 / \underline{65.57} 

\textcolor[RGB]{0,204,102}{$\uparrow$}
& 51.32 / \underline{52.58} 
&   57.55 / \underline{59.98}

\\ 
{WS-MCU} &  \hspace{1.8mm} 53.57 / \underline{57.10}  

\textcolor[RGB]{0,204,102}{$\uparrow$}
& \hspace{1.8mm} 44.48 / \underline{47.16}

\textcolor[RGB]{0,204,102}{$\uparrow$}
&  \hspace{1.8mm} 64.40 / \underline{67.20} 

\textcolor[RGB]{0,204,102}{$\uparrow$}
&   52.12 / \underline{\textbf{53.58}} 

&  \hspace{1.8mm} 56.33 / \underline{59.07} 

\textcolor[RGB]{0,204,102}{$\uparrow$}
\\ 
{WS-EU} &  51.79 / \underline{54.46}  

& \hspace{1.8mm} 42.32 / \underline{46.32} 
\textcolor[RGB]{0,204,102}{$\uparrow$}
&  \hspace{1.8mm} 64.61 / \underline{\textbf{68.68}} 

\textcolor[RGB]{0,204,102}{$\uparrow$}
& 49.80 / \underline{51.81} 
&   55.81 / \underline{57.89}

\\ 
\bottomrule
\end{tabular}
}
% }
\end{table*}

\subsection{Real-World datasets comparisons}\label{sec_rw_comparison}
\noindent\textbf{Datasets and backbones.} Apart from benchmark datasets whose candidate label set needs to be self-generated, here we evaluate our proposed WorseNet-based modules on five real-world datasets that are widely used in PLL: Lost \cite{cour2011learning}, MSRCv2 \cite{liu2012conditional}, BirdSong \cite{briggs2012rank}, Soccer Player \cite{zeng2013learning} and Yahoo!News \cite{guillaumin2010multiple}. Note that all five of these real-world datasets are annotated with the given candidate label sets, and most samples, as a realistic scenario, are annotated with similar semantic labels. Thus, we simply use them as the oracle annotation. For these five datasets, we adopt the same 3-layer MLP used in Section \ref{sec_benchmark_datasets_comparisons} as the sole backbone since these real-world datasets are not limited to image input (simple vector inputs), which also follows conventions in~\cite{feng2019partial,self-training,feng2020provably, lv2020progressive,wen2021leveraged,wang2022pico,zhang2022exploiting}. 

\noindent\textbf{Compared methods and training settings.} Due to the simplicity of these five real-world datasets, we adopt a simple MLP as the backbone for both the predictor and WorseNet, so here we compare our methods with seven model-driven methods, 1) - 7), the architecture of which does not necessarily build upon the deep models. Based on the different data quantities,  we specifically design different settings for these five datasets. Specifically, we set the size of the initial labeled set $\sL$ to $5$, and repeat the query process $5$ times. We repeatedly conduct all experiments 10 times, and record the average testing accuracy by using the model achieving maximum performance on a validation set built by randomly selecting $10$ instances from the training datasets. Other settings are similar to Section~\ref{sec_benchmark_datasets_comparisons}.

\noindent\textbf{Experiment results.} The experimental results in Table~\ref{tab_RW} validate that our proposed WorseNet is also effective in dealing with ALPL in five real-world datasets. Specifically, our WP is capable of delivering promising performance gains to the predictor with any baseline method. Furthermore, the three improved metrics (WS-MMU, WS-MCU, and WS-EU) in the selector also show competitive performance compared to the baselines.

\subsection{Ablation studies on WorseNet}\label{sec_Number of selected samples on WorseNet}
\noindent\textbf{Comparison with different DAs.} In Table~\ref{tab_dataaug}, we implement three \emph{data augmentations} (DAs), i.e., Random Erasing, Mixup (The mixing parameter of MixUp is $0.5$), and Random Crop on top of RS to evaluate the generalization of WorseNet. Clearly, our proposed WorseNet could improve predictor performance with any of three DAs on two evaluated datasets, which further validates the superiority and effectiveness of our proposed WorseNet. We also notice that Random Erasing and Mixup could make harmful performance degradation to the predictor. This may be due to that these DAs further damage the training samples built on the partial labels, especially for Mixup (the partial labels are mixed together), troubling the learning of the predictor in ALPL. In conclusion, our WorseNet is a promising method to address \emph{overfitting}.

\begin{table*}[!t]
\centering
\caption{Testing performance with different data augmentation. The \underline{underline} points out the improved accuracy by WP. Other settings are similar with Table~\ref{tab_bench_FPS_0.5}.}
\label{tab_dataaug}
\resizebox{0.85\textwidth}{!}{
\begin{tabular}{c|c|c}
\toprule
Data Augmentation + RS & Fashion-MNIST & CIFAR-10\\
\midrule
Random Erasing & 66.15 $\pm$ 5.98 / \underline{67.12 $\pm$ 4.25}   & 49.87 $\pm$ 3.22 / \underline{51.02 $\pm$ 2.58}  \\ 
Mixup & 16.15 $\pm$ 4.67 / \underline{16.56 $\pm$ 4.87}  & 10.33 $\pm$ 4.74 / \underline{12.19 $\pm$ 5.23}  \\ 
Random Crop & 71.29 $\pm$ 1.00 / \underline{72.82 $\pm$ 0.51} & 56.33 $\pm$ 1.05 / \underline{57.62 $\pm$ 1.74} 
\\
\bottomrule
\end{tabular}
% }
}
\vspace{-3.5mm}
\end{table*}

\begin{figure*}[!t]
    \centering
    \includegraphics[width=0.90\textwidth]{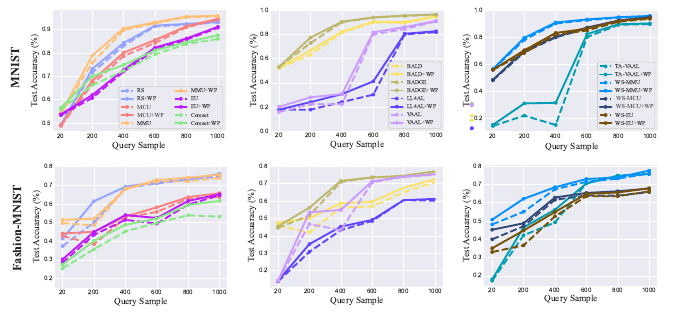}
    \vspace{-3mm}
    \caption{The average test accuracy over the different number of query samples on four benchmark datasets during the training time. Note that here the settings are the same with Table~\ref{tab_bench_FPS_0.5}. }
    \label{fig_fps_0.5}
    % \vspace{-2mm}
\end{figure*}

\noindent\textbf{Number of selected samples.} As shown in Figure \ref{fig_fps_0.5}, with the increase of queried samples ($100$ samples in each round), all methods achieve steady performance enhancement throughout the whole training time. Clearly, it is noticed that all baseline methods (dashed lines) are comparably strengthened by our proposed WP (solid lines) in each query round. Besides, the three new proposed selectors could also achieve competitive performance. More relevant results can be found in the \textbf{Appendix} file.

\section{Conclusion}
We have proposed and investigated a new and practical setting, \emph{active learning with partial labels} (ALPL), where the oracle is requested to provide partial labels for the selected samples during the query process. To address ALPL, we first adopt RC loss on different prevailing AL frameworks to establish a strong and effective baseline. Motivated by the salutary effects of \emph{counter examples} (CEs) in human reasoning, we turn to such a human-based adversarial learning process to relieve the \emph{overfitting} and improve the partially-labeled sample selection process in ALPL. In this regard, we designed CEs by reversing the original partially-labeled examples. Furthermore, we introduced WorseNet that directly learns such complementary knowledge by using the proposed Worse loss. By capitalizing on the probability gap between the predictor and WorseNet, our proposed WorseNet not only explicitly enhances the evaluation performance of the predictor but also improves the selector's ability to query partially-labeled samples more precisely.  Comprehensive experimental results on various
datasets demonstrate that our WorseNet yields state-of-the-art performance in ALPL, and validates the superiority of such an adversarial learning pattern. Additionally, PLL could also be well addressed by this method, which warrants further investigation in the future.

\begin{appendices}

\section{Generation of PLL} \label{sup_geneartion_pll}

In Section 3, we introduce two different generation ways for the candidate label sets, i.e., USS, uniformly sampling a label set from the full partial label space ${\sC}$ for each instance. FPS, setting a flip probability $q$ for any irrelevant labels which could possibly become an item in the candidate label set with probability $q$. 
\subsection{USS}
For USS, each partially-labeled example $(\vx,S)$ is independently drawn from a probability distribution with the following density:
% \begin{equation}\small
% $$\widetilde {P}(\vx,S) = \sum \nolimits_{i=1}^{k}P(S|y=i)P(\vx,y=i), where P(S|y=i) =
% \begin{cases}
% {\frac {1} {2^{k-1}-1}} & \text{i\in S}
% {0}& \text{i \notin S}
% \end{cases}$$

% \end{equation}

\begin{equation}
\widetilde {P}(\vx,S) = \sum \nolimits_{i=1}^{k}P(S|y=i)P(\vx,y=i), P(S|y=i) =
 \begin{cases}
   {\frac {1} {2^{k-1}-1}}  & \text{$i\in S$,} \\
    0   &  \text{$i \notin S$.}\\
 \end{cases}         
 \label{process_i}
 \end{equation}

The generation process assumes that the candidate label set $S$ is independent of the instance $\vx$. There are a total of $2^{k}-1$ possible candidate label sets that contain the specific true label $y$. Therefore, Eq. (\ref{process_i}) illustrates that the candidate label set for each instance is uniformly sampled.

\subsection{FPS}
For FPS, we set a flip probability $q$ to any irrelevant label that possibly entries the candidate label set. Here, we introduce the class transition matrix (denoted by $T$) for partially labeled data, where $T_{ij}$ refers to the probability that the label $j$ is a candidate label given the true label $i$ for each instance. Note that $T_{ii} = 1$ always holds since the true label always belongs to the candidate set. $T_{ij} = q, i \neq j$ holds for other elements. 

\section{Proof and Analysis of Theorem 1}\label{sup_sec_theorem}
Since the WorseNet is regulated to learn the inverse partial-label set $\overline{S}$, the WorseNet shall have a high confidence on the false labels in $\overline{S}$, which is complementary to the predictor. Therefore,
we assume the following equality:
\begin{equation}\label{sup_eq_equavalaent}
Q(y_i=j|\vx_i) + P(y_i=j|\vx_i) = 1, j \in \overline{S}_i, i \in \{1,...,l\}.
\end{equation}

Assume that the loss function $\mathcal{L}$ is implemented with a standard cross entropy loss (which is also the practical achievement in our experiments and ~\cite{feng2020provably}). In this way, we have the following equation for the IRC loss:
\begin{equation}\label{sup_eq_crossentropy}
\mathcal{\hat R}_{\mathrm{irc}} = \sum_{i=1}^{l}\sum_{j \in \overline{S}_i}- \frac{Q(y_i = j|\vx_i)} {\sum_{j \in \overline{S}_i}Q(y_i=j|\vx_i)} \mathrm{log} Q(y_i = j|\vx_i).
\end{equation}

Based on Eq. (\ref{sup_eq_equavalaent}) and (\ref{sup_eq_crossentropy}), we could express the Worse loss $\mathcal{\hat R}_{worse}$ as
\begin{equation}\label{sup_eqa_wloss}
\hspace{-20mm}\begin{aligned}
     \mathcal{\hat R}_{\mathrm{worse}} &= \mathcal{\hat R}_{\mathrm{irc}} (\mathcal{L}, w) +  \alpha \mathrm{KLD} \qquad (\alpha=1)  \\
     & = \sum_{i=1}^{l}\sum_{j \in \overline{S}_i} - \frac{Q(y_i = j|\vx_i)} {\sum_{j \in \overline{S}_i}Q(y_i=j|\vx_i)} \mathrm{log} Q(y_i = j|\vx_i) + \sum_{i=1}^{l} \sum_{j \in \overline{S}_i} {P(y_i = j|\vx_i)} \mathrm{log} \frac {{P(y_i = j|\vx_i)}} {{Q(y_i = j|\vx_i)}} \\
     &= \sum_{i=1}^{l}\sum_{j \in \overline{S}_i} - \frac{Q(y_i = j|\vx_i)} {\sum_{j \in \overline{S}_i}Q(y_i=j|\vx_i)} \mathrm{log} Q(y_i = j|\vx_i) + \underbrace{P(y_i = j|\vx_i) \mathrm{log} P(y_i = j|\vx_i)}_{\text{constant term when minimizing $\mathcal{\hat R}_{\mathrm{worse}}$ }} - P(y_i = j|\vx_i) \mathrm{log} Q(y_i = j|\vx_i)\\
     &= \sum_{i=1}^{l}\sum_{j \in \overline{S}_i} - \frac{Q(y_i = j|\vx_i)} {\sum_{j \in \overline{S}_i}Q(y_i=j|\vx_i)} \mathrm{log} Q(y_i = j|\vx_i) - \underbrace{(1-Q(y_i = j|\vx_i)}_{\text{substitution for $P(y_i = j|\vx_i)$) by Eq. (\ref{sup_eq_equavalaent}) }} \mathrm{log} Q(y_i = j|\vx_i) + c\\
     &= \sum_{i=1}^{l}\sum_{j \in \overline{S}_i} - \frac{Q(y_i = j|\vx_i)} {\sum_{j \in \overline{S}_i}Q(y_i=j|\vx_i)} \mathrm{log} Q(y_i = j|\vx_i) - (1-Q(y_i = j|\vx_i)\mathrm{log} Q(y_i = j|\vx_i) + c\\
     &= \sum_{i=1}^{l}\sum_{j \in \overline{S}_i}- \underbrace{\frac{Q(y_i = j|\vx_i)+(1-Q(y_i = j|\vx_i))\sum_{j \in \overline{S}_i}Q(y_i=j|\vx_i)} {\sum_{j \in \overline{S}_i}Q(y_i=j|\vx_i)}}_{n(Q(y_i=j|\vx_i))}\mathrm{log}Q(y_i = j|\vx_i) + c\\
     &= \sum_{i=1}^{l}\sum_{j \in \overline{S}_i}-n(Q(y_i=j|\vx_i)\mathrm{log}Q(y_i = j|\vx_i) + c.
\end{aligned}
\end{equation}

Intuitively, minimizing the Worse loss $\mathcal{\hat R}_{\mathrm{worse}}$ is equal to minimize the loss function below:
\begin{equation}\label{sup_final_eqa}
 \min \mathcal{\hat R}_{\mathrm{worse}} \Longrightarrow \min \sum_{i=1}^{l}\sum_{j \in \overline{S}_i}-n(Q(y_i=j|\vx_i))\mathrm{log}Q(y_i = j|\vx_i), Q(y_i = j|\vx_i) \in [0,1].
\end{equation}

\begin{figure*}[h]
\centering
\includegraphics[width=0.5\textwidth]{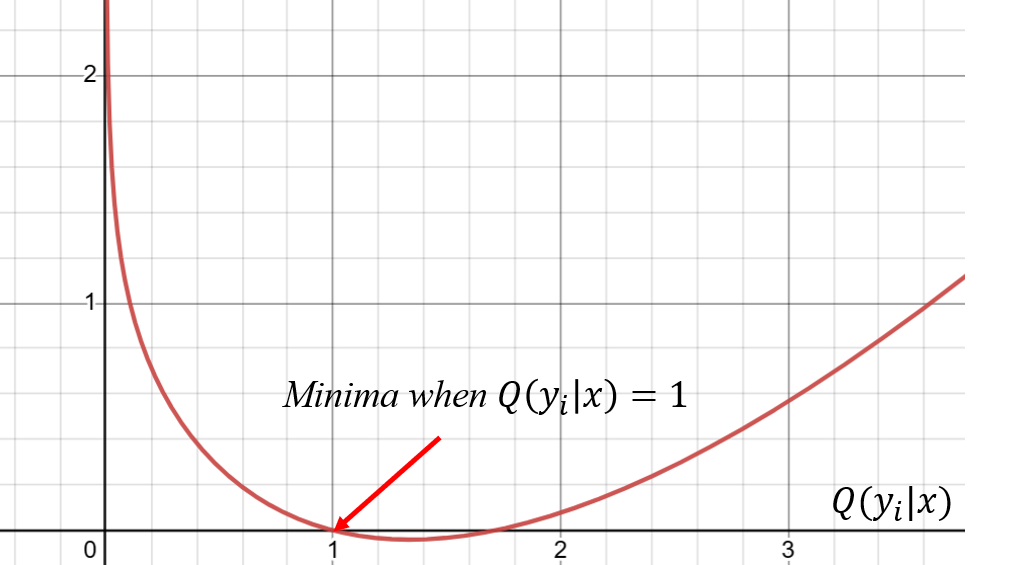}
\caption{The loss graph of the simplified Worse 
 loss $ \mathcal{\hat R}_{\mathrm{worse}}$ in Eq. (\ref{sup_final_eqa}).}
 
\label{sup_fig_loss}
\end{figure*}

Here, the proof of Theorem 1 is complete. Figure~\ref{sup_fig_loss} illustrates the graph of the above function, from which we can easily observe that this loss is a monotone-decreasing function. While optimizing the Worse loss, the probability of WorseNet $Q(y_i=j|\vx_i)$ gradually approaches ``1", indicating high confidence in the false labels in the inverse partial-label set $\overline{S}$. Since learning the WorseNet is complementary to the predictor, the predictor is expected to decrease its confidence in predicting labels in $\overline{S}$ ($P(y_i = j|\vx_i) \rightarrow 0$). In other words, WorseNet acts as an auxiliary regularization to the RC loss, further keeping the predictor away from the false labels in $\overline{S}$. We believe this is also why this learning mechanism can help the predictor alleviate \emph{overfitting} problems, as the predictor is able to generalize to the entire label space (whereas RC loss only considers the labels in the partial label set $S$).

\section{ Experiments}\label{appendix_experiments}
\subsection{Benchmark Datasets}
In Section 5.1, we use four widely-used benchmark datasets, i.e.,
 MNIST~\cite{minist}, Fashion-MNIST~\cite{fashion}, SVHN~\cite{svhn}, and CIFAR-10~\cite{cifar}. Table~\ref{chara1} lists the characteristics of these datasets. We respectively describe these datasets as follows.
\begin{itemize}
\item MNIST~\cite{minist}: It is a 10-class dataset of handwritten digits. Each data is a 28 × 28 grayscale image.
\item Fashion-MNIST~\cite{fashion}: It is also a 10-class dataset. Each instance
is a fashion item from one of the 10 classes, which are T-shirt/top, trouser, pullover, dress, sandal, coat, shirt, sneaker, bag, and ankle boot. Moreover, each image is a 28 × 28 grayscale image.
\item SVHN~\cite{svhn}: Each instance is a 32 × 32 × 3 colored
image in RGB format. It is a 10-class dataset of digits.
\item CIFAR-10~\cite{cifar}: Each instance is a 32 × 32 × 3 colored image
in RGB format. It is a ten-class dataset of objects including
airplane, bird, automobile, cat, deer, frog, dog, horse, ship,
and truck.
\end{itemize}
\begin{table}[!h]
\caption{Characteristics of benchmark datasets}
\centering
\begin{tabular}{c|c|c|c|c}
\toprule
Datasets & \#Train & \#Test & \#Features  & \#Classes \\
\midrule
MNIST~\cite{minist} & 60,000 & 10,000 & 784 & 10\\
Fashion-MNIST~\cite{fashion} & 60,000 & 10,000 & 784 & 10\\
SVHN~\cite{svhn} & 73,257 & 26,032 & 3,072 & 10\\
CIFAR-10~\cite{cifar} & 50,000 & 10,000 & 3,072 & 10\\
\bottomrule
\end{tabular}
\label{chara1}
\end{table}

\subsection{Real Datasets}\label{appendix_real}
In Section 5.2, we select five real-world datasets including Lost~\cite{cour2011learning}, MSRCv2~\cite{liu2012conditional}, BirdSong~\cite{briggs2012rank}, Soccer Player~\cite{zeng2013learning}, and Yahoo!News~\cite{guillaumin2010multiple}. According to the different data quantities of these datasets, we specifically design the unique query setting for each of them, and the detailed parameters are in Table~\ref{tab_RW_setting}. Here, we give a comprehensive description of them as follows.

\begin{itemize}

\item Lost, Soccer Player, and Yahoo!News: They crop faces in
images or video frames as instances, and the names appearing
on the corresponding captions or subtitles are considered as
candidate labels.
\item MSRCv2: Each image segment is treated as a sample, and objects appearing in the same image are regarded as candidate
labels.
\item BirdSong: The singing syllables of birds are regarded as instances,
and bird species that are jointly singing during any ten seconds are represented as candidate labels.
\end{itemize}

\begin{table}[!htbp]
\caption{Characteristics of the real-world datasets.}
\centering
\begin{tabular}{c|c|c|c|c|c}
\toprule
Datasets & Application Domain&\#Examples & \#Features  & \#Classes &Avg \#CLs \\
\midrule
Lost~\cite{cour2011learning} & Automatic face naming & 1,122 & 108 & 16 & 2.23\\
MSRCv2~\cite{liu2012conditional} & Object classification & 1,758 & 48 & 23&3.16\\
BirdSong~\cite{briggs2012rank} & Bird song classification & 4,998 & 38 & 13 & 2.18\\
Soccer Player~\cite{zeng2013learning} & Automatic face naming & 17,472 & 279 & 171 & 2.09\\
Yahoo! News~\cite{guillaumin2010multiple}& Automatic face naming & 22,991 & 163 & 219 & 1.91\\
\bottomrule
\end{tabular}
\label{chara2}
\end{table}

\begin{table}[!htbp]
\centering
\caption{The explicit query size $b$ and budget size $B$ on five real-world datasets in ALPL. The percentage number(\%) depicts the proportion of query budget in the total unlabeled data.}
\label{tab_RW_setting}
% \resizebox{1.0\textwidth}{!}{
% \setlength{\tabcolsep}{1mm}{
\begin{tabular}{c|c|c}
\toprule
Parameters & {Query size ($b$)}  & {Query budget ($B$)} \\
\midrule
{Lost}~\cite{cour2011learning} & 40 & 200 (17.8\%) \\
MSRCV2~\cite{liu2012conditional} & 60 &  300  (17.1\%) \\
BirdSong~\cite{briggs2012rank} & 200 & 1000  (20.0\%)\\
SoccerPlayer~\cite{zeng2013learning} & 600  & 3000  (17.2\%) \\
Yahoo!News~\cite{guillaumin2010multiple} & 900  & 4500  (19.6\%)\\  
\bottomrule
\end{tabular}
% }
% }
\vspace{-6mm}
\end{table}

\subsection{Compared Methods}
In this section we will briefly introduce ten compared methods used in Section 5.5, containing seven model-based modules and three data-driven modules. The compared methods are list as follows:

\begin{itemize}

\item[1)] Random Sampling (RS): In each query round, it randomly selects $b$ samples from the unlabeled pool, and then hand over these samples to the oracle for annotation.

\item[2)] Minimum confidence uncertainty (MCU): Similar to MMU, it calculates the uncertainty score but using Eq. 1 and selects the $b$ samples with the highest uncertainty scores in the unlabeled pool and then sends these samples to the oracle for annotation.

\item[3)] Minimum margin uncertainty (MMU): In each query round, it calculates the uncertainty score using Eq. 2 and selects the $b$ samples with the highest uncertainty scores in the unlabeled pool and then sends these samples to the oracle for annotation.

\item[4)] Entropy uncertainty (EU): Similar to MMU, it uses Eq. 3 to obtain the uncertainty score in each round, and selects the $b$ samples with the highest uncertainty scores in the unlabeled pool and then sends these samples to the oracle for annotation.

\item[5)] Coreset~\cite{coreset}: In each query round, it selects $b$ samples by solving a $b$-center issues on the full unlabeled space, using the embedding of the unlabeled samples generated from the penultimate layer of the predictor. 

\item[6)] BALD~\cite{batchbald}: It is developed based on~\cite{bald}. The original version~\cite{bald} is a Bayesian modelling-based method, combining the Bayesian modelling to calculate the uncertainty score in each query round. BALD improves this mechanism and proposes an acquisition
function to select multiple informative points jointly for AL.

\item[7)] BADGE~\cite{badge}: It selects $b$ samples by adopting the $k-$Means++ to group the features in the unlabeled space, and the feature is generated in a hallucinated gradient space.

\item[8)] LL4AL~\cite{lloss}: It introduces an extra module to learn the loss of the predictor, and selects $b$ samples by the loss distance between the predictor and the extra module, and then hands these samples to the oracle for annotation.

\item[9)] VAAL~\cite{vaal}: It proposes to train a VAE, latching on to the representing information of both the labeled and unlabeled data. With the help of adversarial learning, the selector could choose $b$ samples with high diversity compared to the labeled samples. 

\item[10)] TA-VAAL~\cite{tvaal}: Building upon VAAL, it further exploits the space difference between the labeled data and the unlabeled data, and incorporate the "learning loss"~\cite{lloss} module to select better representative samples in each query round.
\end{itemize}

\subsection{Ablation results on WorseNet}\label{appendix_Ablation results on WorseNet-Predictor}

\noindent\textbf{Different Backbones and partial label generation approaches.} In Section 5.1, we list the test performance of our proposed Worsenet and ten AL-based approaches with C-Net (ResNet18) for MINIST and Fashion-MNIST (SVHN and CIFAR-10), and the partial labels are generated using FPS ($q = 0.5$). Here we show the corresponding results implemented based on different backbones and partial label generation methods among Tables~\ref{tab_bench_uss}-\ref{tab_bench_otherback_fps_0.5}. As shown in these tables, we could tell that our proposed WP achieves global improvements on all proposed AL-based methods among all backbones and partial label generation methods. Specifically, our proposed WP could achieve performance elation in both FPS with $q=0.3$ and $q=0.5$ cases, illustrating that WP is robust to the label noise in the candidate set.

\subsection{Ablation results on WorseNet}\label{appendix_Ablation results on WorseNet-Predictor}

\noindent\textbf{Different Backbones and partial label generation approaches.} In Section 5.1, we list the test performance of our proposed Worsenet and ten AL-based approaches with C-Net (ResNet18) for MINIST and Fashion-MNIST (SVHN and CIFAR-10), and the partial labels are generated using FPS ($q = 0.5$). Here we show the corresponding results implemented based on different backbones and partial label generation methods among Tables~\ref{tab_bench_uss}-\ref{tab_bench_otherback_fps_0.5}. As shown in these tables, we could tell that our proposed WP achieves global improvements on all proposed AL-based methods among all backbones and partial label generation methods. Specifically, our proposed WP could achieve performance elation in both FPS with $q=0.3$ and $q=0.5$ cases, illustrating that WP is robust to the label noise in the candidate set.

\noindent\textbf{Discussion about WorseNet-Selector module.}
For the three newly designed uncertainty-based selectors, i.e., WS-MMU, WS-MCU, and WS-EU, it is found that they could achieve a much higher performance gain in some cases compared to the original version. For instance, WS-MMU achieves about 18\% accuracy elation compared to MMU in Table~\ref{tab_bench_otherback_fps_0.5}. However, it is admitted that WS sometimes degrades the original selection strategies. As shown in Table~\ref{tab_bench_otherback_fps_0.3}, we can see that WS-MCU are inferior (about 1\% accuracy decline) to MCU in Fashion-MINST. More similar phenomenon inordinately appears in different situations in Tables~\ref{tab_bench_uss}-\ref{tab_bench_otherback_fps_0.5}.

\subsection{Ablation studies on the number of selected samples on WorseNet}\label{appendix_Ablation studies on the number of selected samples on WorseNet}

\begin{figure}[!htbp]
    \centering
    \includegraphics[width=1.0\textwidth]{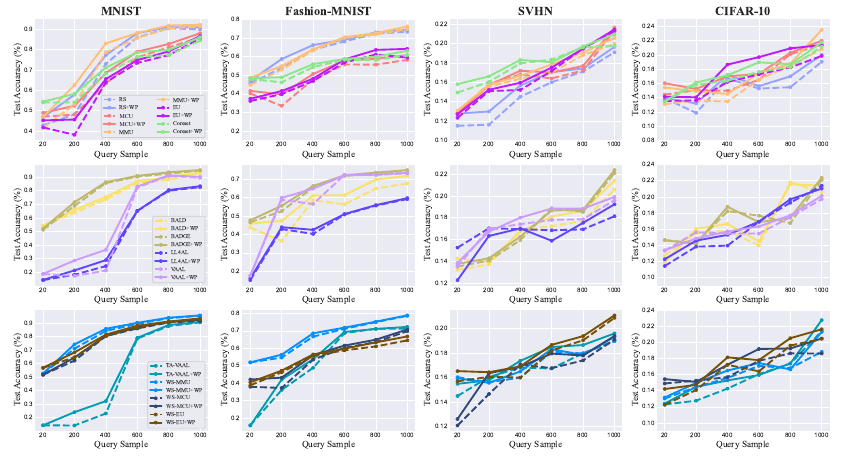}
    \caption{The average test accuracy over the different number of query samples on four benchmark datasets during the training time. Note that here settings are corresponding to Table~\ref{tab_bench_uss} (USS).}
    \label{fig_uss}
\end{figure}

\begin{figure}[!htbp]
    \centering
\includegraphics[width=1.0\textwidth]{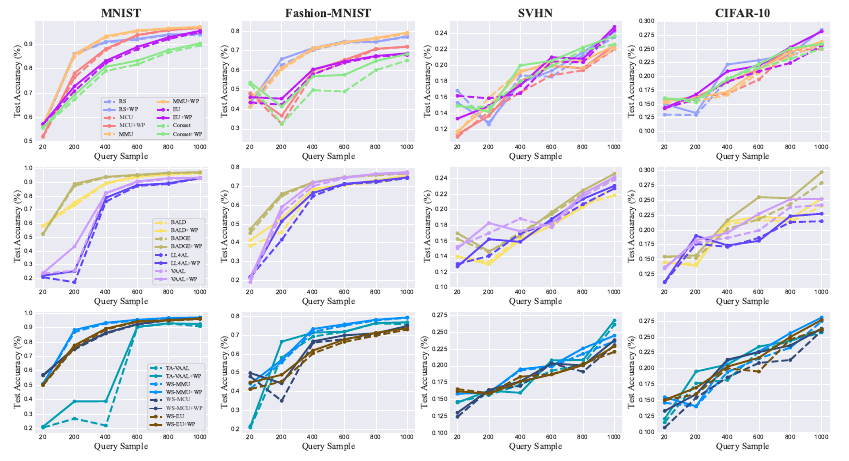}
    \caption{The average test accuracy over the different number of query samples on four benchmark datasets during the training time. Note that here settings are corresponding to Table~\ref{tab_bench_FPS_0.3} (FPS with $q=0.3$).}
    \label{fig_fps_0.3}
\end{figure}

In Section 5.3, we study the influence of the number of selected samples during the training period over all modules. Here we present more relevant results in different cases. Figure~\ref{fig_fps_0.3} (Figure~\ref{fig_uss}) shows the results in FPS with $q=0.3$ (USS), and we can find that our proposed WP (solid lines) could achieve sustainable improvements in all baseline methods (dashed lines) regardless of the partial label generation approach. Besides, we can find that the enhancements are not obvious for some data-driven methods such as LL4AL and VAAL, which means our proposed WP module could be further refined.  
 
\begin{table*}[!htbp]
\centering
\caption{Test performance of the proposed WorseNet modules and other methods on benchmark datasets using label generation by USS. The best results among all methods with the same backbone are marked in \textbf{bold}. -/+ WP denotes whether the predictor is helped by WorseNet. The \underline{underline} points out improved accuracy by WP. \textcolor[RGB]{0,204,102}{$\uparrow$} indicates the improved accuracy is beyond 1\%. The backbones for MNIST and Fashion-MINIST are C-Net, and for SVHN and CIFAR-10 are ResNet18. Here the standard deviation is ignored. }
\label{tab_bench_uss}
\resizebox{1.0\textwidth}{!}{
\setlength{\tabcolsep}{4mm}{
\begin{tabular}{c|c|c|c|c}
\toprule
% \makebox[0.1\textwidth][c]{name} & \makebox[0.2\textwidth][c]{taskA} & 
%                                      & \makebox[0.4\textwidth][c]{taskC} & \makebox[0.2\textwidth][c]{taskD}
Methods ( -/+ WP) & MNIST & Fashion-MINIST & SVHN & CIFAR-10 \\
\midrule
{RS} & 90.95 / \underline{91.82}
&  74.05  / \underline{74.66} 
& 79.19  / \underline{79.64} 
& \hspace{1.8mm} 56.91 / \underline{58.62}

\textcolor[RGB]{0,204,102}{$\uparrow$}\\   

{MMU} & \hspace{1.8mm} 92.66 / \underline{93.90}

\textcolor[RGB]{0,204,102}{$\uparrow$} 
& \hspace{1.8mm} 74.80 / \underline{76.19}  

\textcolor[RGB]{0,204,102}{$\uparrow$} 
& \hspace{1.8mm} 80.14 / \underline{81.73} 

\textcolor[RGB]{0,204,102}{$\uparrow$} 
& \hspace{1.8mm} 56.77 / \underline{\textbf{59.54}} 

\textcolor[RGB]{0,204,102}{$\uparrow$} 
\\    

{MCU} & \hspace{1.8mm} 85.91 / \underline{87.97} 

\textcolor[RGB]{0,204,102}{$\uparrow$}
&  \hspace{1.8mm} 59.64 / \underline{61.60} 

\textcolor[RGB]{0,204,102}{$\uparrow$}
& 79.88 / \underline{80.44} 
& 57.45 / \underline{57.99} \\

{EU} & \hspace{1.8mm} 85.33 / \underline{86.59}

\textcolor[RGB]{0,204,102}{$\uparrow$}
&  \hspace{1.8mm} 62.36 / \underline{64.73} 

\textcolor[RGB]{0,204,102}{$\uparrow$}
& 79.97 / \underline{81.06}
& \hspace{1.8mm} 58.15 / \underline{60.46} 

\textcolor[RGB]{0,204,102}{$\uparrow$} \\    

{Coreset} & \hspace{1.8mm} 84.33 / \underline{86.10} 

\textcolor[RGB]{0,204,102}{$\uparrow$}
& \hspace{1.8mm} 64.34 / \underline{65.79} 

\textcolor[RGB]{0,204,102}{$\uparrow$}
& \hspace{1.8mm} 79.73 / \underline{81.05}

\textcolor[RGB]{0,204,102}{$\uparrow$}
& 55.43 / \underline{58.05} \\    

{BALD} & 93.50 / \underline{93.90}
& \hspace{1.8mm} 69.55 / \underline{72.68}

\textcolor[RGB]{0,204,102}{$\uparrow$}
& 80.48 / \underline{81.39}
& 57.80 / \underline{58.17}  \\  

{BADGE} & 95.00 / \underline{95.25}
& 74.82 / \underline{75.75} 

& 82.09  / \underline{\textbf{82.47}}  
& 58.18  / \underline{58.58}\\  

{LL4AL} & 82.74 / \underline{83.31}

& 59.10 / \underline{59.65}
& 79.73  / \underline{79.94} 
& \hspace{1.8mm} 57.02 / \underline{58.87}  

\textcolor[RGB]{0,204,102}{$\uparrow$}\\ 

{VAAL} & \hspace{1.8mm} 90.98 / \underline{91.21} 

\textcolor[RGB]{0,204,102}{$\uparrow$}
& 73.12 / \underline{73.83}

& 79.11 / \underline{79.75}
& 57.72 / \underline{58.12}
\\ 

{TA-VAAL} & 90.85 / \underline{91.13}
& 71.94 / \underline{72.45} 
& 79.07 / \underline{80.33}  
& 58.14 / \underline{58.73}   \\ 
\midrule
{WS-MMU} & 95.21 / \underline{\textbf{95.54}}
& 78.55 / \underline{\textbf{78.80}} 
& 80.06 / \underline{80.78} 
& \hspace{1.8mm} 57.38 / \underline{58.48}  

\textcolor[RGB]{0,204,102}{$\uparrow$}\\ 
{WS-MCU} & 92.44 / \underline{92.90}
& 70.52 / \underline{71.50}
& 81.10 / \underline{81.39} 
& 57.17 / \underline{59.14}   \\ 
{WS-EU} & 93.56 / \underline{94.03} 
& \hspace{1.8mm} 65.62  / \underline{67.99} 

\textcolor[RGB]{0,204,102}{$\uparrow$}
& 80.87 / \underline{81.07} 
& \hspace{1.8mm} 58.36 / \underline{60.36}  

\textcolor[RGB]{0,204,102}{$\uparrow$}\\ 
\bottomrule
\end{tabular}
}
}
\end{table*}

\begin{table*}[!htbp]
\centering
\caption{Test performance of the proposed WorseNet modules and other methods on benchmark datasets using label generation  by FPS ($q = 0.3$). The best results among all methods with the same backbone are marked in \textbf{bold}. -/+ WP denotes whether the predictor is helped by WorseNet. The \underline{underline} points out improved accuracy by WP. \textcolor[RGB]{0,204,102}{$\uparrow$} indicates the improved accuracy is beyond 1\%. The backbones for MNIST and Fashion-MINIST are C-Net, and for SVHN and CIFAR-10 are ResNet18. Here the standard deviation is ignored. }
\label{tab_bench_FPS_0.3}
\resizebox{1.0\textwidth}{!}{
\setlength{\tabcolsep}{4mm}{
\begin{tabular}{c|c|c|c|c}
\toprule
Methods ( -/+ WP) & MNIST & Fashion-MINIST & SVHN & CIFAR-10 \\
\midrule
{RS} & 94.18 / \underline{94.51}
& 77.53 / \underline{77.82} 
& 80.52 / \underline{81.19} 
& \hspace{1.8mm} 58.83 / \underline{61.46} \textcolor[RGB]{0,204,102}{$\uparrow$} \\   

{MMU} & 96.99 / \underline{97.21} 
&  79.35 / \underline{79.44} 
& 82.10 / \underline{82.61} 
& 60.96 / \underline{61.85}  \\ 

{MCU} & 96.65 / \underline{96.76} 
&  72.16 / \underline{72.35} 
& 82.05 / \underline{82.46}
& 62.84 / \underline{63.31} \\

{EU} & 94.84 / \underline{95.41} 
& \hspace{1.8mm}  68.99 / \underline{70.51}

\textcolor[RGB]{0,204,102}{$\uparrow$}
& 84.40 / \underline{84.79} 
& \hspace{1.8mm} 61.34 / \underline{62.17} 

\textcolor[RGB]{0,204,102}{$\uparrow$}\\    

{Coreset} & 89.71 / \underline{90.76} 
& \hspace{1.8mm}  64.98 / \underline{68.26}
 \textcolor[RGB]{0,204,102}{$\uparrow$}
& \hspace{1.8mm}  82.65 / \underline{83.65} 
\textcolor[RGB]{0,204,102}{$\uparrow$}
& 61.02 / \underline{62.88}   \\    

{BALD} & 96.61 / \underline{96.74} 
&  75.59 / \underline{75.84} 
& \hspace{1.8mm} 81.82 / \underline{82.85} 

\textcolor[RGB]{0,204,102}{$\uparrow$}
& \hspace{1.8mm}  60.12 / \underline{61.35}  

\textcolor[RGB]{0,204,102}{$\uparrow$}
\\  

{BADGE} & 97.08 / \underline{\textbf{97.37}} 
& 77.86 / \underline{78.30} 
& 83.88 / \underline{84.61}
& 63.88 / \underline{\textbf{64.69}}   \\  

{LL4AL} & 92.85 / \underline{93.11} 
& 75.09 / \underline{75.58} 
& 82.75 / \underline{83.15} 
& \hspace{1.8mm}  57.44 / \underline{58.79}

\textcolor[RGB]{0,204,102}{$\uparrow$}\\ 

{VAAL} & 93.36 / \underline{93.61}
& 77.72 / \underline{77.98} 
& 83.83 / \underline{84.19} 
& \hspace{1.8mm}  60.15 / \underline{61.16} 

\textcolor[RGB]{0,204,102}{$\uparrow$}  \\ 

{TA-VAAL} & 93.07 / \underline{93.30} 
& 76.94 / \underline{77.44} 
& 86.11 / \underline{\textbf{86.74}} 
& 61.69 / \underline{62.19}   \\ 
\midrule
{WS-MMU} & 97.11 / \underline{97.35} 
& 79.47 / \underline{\textbf{79.80}} 
& 83.66 / \underline{84.49}
& 63.46 / \underline{64.07}   \\ 
{WS-MCU} & 96.15 / \underline{96.41}
& 74.96 / \underline{75.28} 
& 83.08 / \underline{83.81} 
& 61.98 / \underline{63.32}   \\ 
{WS-EU} & 96.10 / \underline{96.33} 
& 74.01 / \underline{74.51}
& 84.08 / \underline{84.91}
& \hspace{1.8mm}  62.16 / \underline{63.69}  
\textcolor[RGB]{0,204,102}{$\uparrow$}
\\ 
\bottomrule
\end{tabular}
}
}

\end{table*}

\begin{table*}[!htbp]
\centering
\caption{Test performance of the proposed WorseNet modules and other methods on benchmark datasets using label generation  by USS. The best results among all methods with the same backbone are marked in \textbf{bold}. -/+ WP denotes whether the predictor is helped by WorseNet. The \underline{underline} points out improved accuracy by WP. \textcolor[RGB]{0,204,102}{$\uparrow$} indicates the improved accuracy is beyond 1\%. The backbones for MNIST and Fashion-MINIST are MLP, and for SVHN and CIFAR-10 are VGG11. Here the standard deviation is ignored. }
\label{tab_bench_otherback_uss}
\resizebox{1.0\textwidth}{!}{
\setlength{\tabcolsep}{4mm}{
\begin{tabular}{c|c|c|c|c}
\toprule
Methods ( -/+ WP) & MNIST & Fashion-MINIST & SVHN & CIFAR-10 \\
\midrule
{RS} & 84.71 / \underline{85.05} 
&  76.32 / \underline{76.76} 
& \hspace{1.8mm}  83.15 / \underline{84.35} 

\textcolor[RGB]{0,204,102}{$\uparrow$}
& 59.64 / \underline{60.19}  \\   

{MMU} & 85.76 / \underline{86.03} 
&  77.76 / \underline{78.24}  
& \hspace{1.8mm}  84.70 / \underline{86.40}  

\textcolor[RGB]{0,204,102}{$\uparrow$}
& 62.89 / \underline{63.17}  \\ 

{MCU} & 77.77 / \underline{78.23} 
&  68.53 / \underline{69.04}  
&  \hspace{1.8mm} 84.31 / \underline{86.01} 

\textcolor[RGB]{0,204,102}{$\uparrow$}
& 61.45 / \underline{62.66}  \\

{EU} & 78.36 / \underline{78.81} 
&  63.70 / \underline{64.52} 
&  \hspace{1.8mm} 84.38 / \underline{86.88} 

\textcolor[RGB]{0,204,102}{$\uparrow$}
& 61.53 / \underline{61.97} \\    

{Coreset} & 70.73 / \underline{71.58} 
&  67.18 / \underline{67.86} 
& 85.57 / \underline{86.25} 
& 60.66 / \underline{60.98}   \\    

{BALD} & 67.18 / \underline{67.56} 
&  73.52 / \underline{74.25}   
&  \hspace{1.8mm} 87.07 / \underline{\textbf{88.37}} 

\textcolor[RGB]{0,204,102}{$\uparrow$}
& 61.88 / \underline{62.22}   \\  

{BADGE} & 86.37 / \underline{86.90}  
& 76.82 / \underline{77.36} 
& 86.18 / \underline{87.19} 
& 63.59 / \underline{\textbf{64.05}}     \\  

\midrule
{WS-MMU} & 87.94 / \underline{\textbf{88.03}} 
& 78.45 / \underline{\textbf{78.97}} 
& \hspace{1.8mm}  86.98 / \underline{87.80} 

\textcolor[RGB]{0,204,102}{$\uparrow$}
& 61.67 / \underline{62.01}    \\ 
{WS-MCU} & 82.38 / \underline{82.67}  
& 72.61 / \underline{73.12} 
&  \hspace{1.8mm} 86.17 / \underline{87.32} 

\textcolor[RGB]{0,204,102}{$\uparrow$}
& 61.84 / \underline{62.14}    \\ 
{WS-EU} & 83.18 / \underline{83.40} 
& 67.85 / \underline{68.23}  

& 86.45 / \underline{87.29} 
& 61.42 / \underline{61.92}   \\ 
\bottomrule
\end{tabular}
}
}
\end{table*}

\begin{table*}[!htbp]
\centering
\caption{Test performance of the proposed WorseNet modules and other methods on benchmark datasets using label generation  by FPS ($q = 0.3$). The best results among all methods with the same backbone are marked in \textbf{bold}. -/+ WP denotes whether the predictor is helped by WorseNet. The \underline{underline} points out improved accuracy by WP. \textcolor[RGB]{0,204,102}{$\uparrow$} indicates the improved accuracy is beyond 1\%. The backbones for MNIST and Fashion-MINIST are MLP, and for SVHN and CIFAR-10 are VGG11. Here the standard deviation is ignored. }
\label{tab_bench_otherback_fps_0.3}
\resizebox{1.0\textwidth}{!}{
\setlength{\tabcolsep}{4mm}{
\begin{tabular}{c|c|c|c|c}
\toprule
Methods ( -/+ WP) & MNIST & Fashion-MINIST & SVHN & CIFAR-10 \\
\midrule
{RS} & 87.30 / \underline{87.96} 
&  79.13 / \underline{79.63} 
&  \hspace{1.8mm} 82.19 / \underline{84.93} 

\textcolor[RGB]{0,204,102}{$\uparrow$}
&  \hspace{1.8mm} 57.63 / \underline{58.98}   

\textcolor[RGB]{0,204,102}{$\uparrow$}\\

{MMU} & 91.44 / \underline{\textbf{91.91}} 
&  80.14 / \underline{\textbf{80.83}}  
&  \hspace{1.8mm} 85.06 / \underline{86.36} 

\textcolor[RGB]{0,204,102}{$\uparrow$}
&  \hspace{1.8mm} 63.19 / \underline{64.49}  

\textcolor[RGB]{0,204,102}{$\uparrow$}
\\    

{MCU} & 88.60 / \underline{89.12} 
&  73.66 / \underline{74.12} 
& 82.34 / \underline{83.37} 

& 64.73 / \underline{65.14}  \\ 
{EU} & 85.64 / \underline{86.41}

&  66.79 / \underline{67.53}

& 86.82 / \underline{87.14}

& 64.69 / \underline{65.08}  

\\

{Coreset} & 73.47 / \underline{74.24}

&  65.75 / \underline{66.21}  
&  \hspace{1.8mm} 85.86 / \underline{87.01} 

\textcolor[RGB]{0,204,102}{$\uparrow$}
&64.93 / \underline{65.33}    \\    

{BALD} & 90.19 / \underline{90.80} 
&  79.10 / \underline{79.68}   
&  \hspace{1.8mm} 87.55 / \underline{\textbf{89.35}} 

\textcolor[RGB]{0,204,102}{$\uparrow$}
&  \hspace{1.8mm} 64.61 / \underline{65.91} 
\textcolor[RGB]{0,204,102}{$\uparrow$}
\\  

{BADGE} & 91.09 / \underline{91.41} 
& \hspace{1.8mm}  78.09 / \underline{79.91} 

\textcolor[RGB]{0,204,102}{$\uparrow$}
& \hspace{1.8mm}  86.94 / \underline{88.46}

\textcolor[RGB]{0,204,102}{$\uparrow$}
&  \hspace{1.8mm} 65.97 / \underline{66.43}     

\textcolor[RGB]{0,204,102}{$\uparrow$}
\\  

\midrule
{WS-MMU} & 91.08 / \underline{91.48} 
& 78.84 / \underline{79.42}

& 84.13 / \underline{84.44} 
& 64.42 / \underline{\textbf{64.74}}    \\ 
{WS-MCU} & 89.14 / \underline{89.98} 
& 72.47 / \underline{73.12} 
& 86.25 / \underline{88.17} 
& 61.74 / \underline{61.81}   \\ 
{WS-EU} & 88.11 / \underline{88.98} 

& 74.07 / \underline{74.59} 
& 87.93 / \underline{88.30}  
& 63.13 / \underline{63.95}   \\ 
\bottomrule
\end{tabular}
}
}
\end{table*}

\begin{table*}[!htbp]
\centering
\caption{Test performance of the proposed WorseNet modules and other methods on benchmark datasets using label generation  by FPS ($q = 0.5$). The best results among all methods with the same backbone are marked in \textbf{bold}. -/+ WP denotes whether the predictor is helped by WorseNet. The \underline{underline} points out improved accuracy by WP. \textcolor[RGB]{0,204,102}{$\uparrow$} indicates the improved accuracy is beyond 1\%. The backbones for MNIST and Fashion-MINIST are MLP, and for SVHN and CIFAR-10 are VGG11. Here the standard deviation is ignored. }

\label{tab_bench_otherback_fps_0.5}
\resizebox{1.0\textwidth}{!}{
\setlength{\tabcolsep}{4mm}{
\begin{tabular}{c|c|c|c|c}
\toprule
Methods ( -/+ WP) & MNIST & Fashion-MINIST & SVHN & CIFAR-10 \\
\midrule
{RS} & 85.45 / \underline{86.17} 
& 77.18 / \underline{77.85}  
&  \hspace{1.8mm} 73.45 / \underline{76.07} 

\textcolor[RGB]{0,204,102}{$\uparrow$}
& 52.91 / \underline{56.21}  \\

{MMU} & 89.14 / \underline{\textbf{89.48}} 
&  78.14 / \underline{\textbf{78.89}}
& \hspace{1.8mm}  74.86 / \underline{75.95}  

\textcolor[RGB]{0,204,102}{$\uparrow$}
& 58.04 / \underline{58.17}   \\    

{MCU} & 82.13 / \underline{82.94}  
&  59.16 / \underline{59.69} 
& 75.15 / \underline{75.48} 
& 59.11 / \underline{59.91}   \\  

{EU} & 79.74 / \underline{80.06} 

&  64.97 / \underline{65.02}  
& 74.25 / \underline{74.53} 
& 58.03 / \underline{58.71}   \\

{Coreset} & 72.18 / \underline{72.59} 
&  \hspace{1.8mm}  63.52 / \underline{64.85}  

\textcolor[RGB]{0,204,102}{$\uparrow$}
&  \hspace{1.8mm} 78.66 / \underline{80.09} 

\textcolor[RGB]{0,204,102}{$\uparrow$}
& \hspace{1.8mm}  60.08 / \underline{61.09}   

\textcolor[RGB]{0,204,102}{$\uparrow$}
\\    

{BALD} & 87.40 / \underline{87.67} 
&  74.67 / \underline{75.58} 

& \hspace{1.8mm}  77.68 / \underline{78.91}

\textcolor[RGB]{0,204,102}{$\uparrow$}
& 61.79 / \underline{62.47}

\\  

{BADGE} & 88.91 / \underline{89.12} 
& 76.97 / \underline{77.19} 
& \hspace{1.8mm}  78.14 / \underline{79.87} 

\textcolor[RGB]{0,204,102}{$\uparrow$}
& 61.54 / \underline{\textbf{62.50}}      \\  

\midrule
{WS-MMU} &  \hspace{1.8mm} 88.45 / \underline{88.56} 

\textcolor[RGB]{0,204,102}{$\uparrow$}
& 77.34 / \underline{78.19}

& 78.53 / \underline{79.07} 
& 59.95 / \underline{60.14}    \\ 
{WS-MCU} & 85.10 / \underline{85.40} 
& \hspace{1.8mm}  70.13 / \underline{71.69} 

\textcolor[RGB]{0,204,102}{$\uparrow$}
& 80.69 / \underline{\textbf{81.45}} 
& 60.73 / \underline{61.06}    \\ 
{WS-EU} & \hspace{1.8mm}  82.80 / \underline{83.91} 

\textcolor[RGB]{0,204,102}{$\uparrow$}
& 66.36 / \underline{66.73} 
& 77.18 / \underline{77.58} 
&\hspace{1.8mm} 59.34 / \underline{60.97}  
\textcolor[RGB]{0,204,102}{$\uparrow$}
\\ 
\bottomrule
\end{tabular}
}
}
\end{table*}

\end{appendices}

\bibliography{sn-bibliography}% common bib file


\begin{thebibliography}{63}
\ifx \bisbn   \undefined \def \bisbn  #1{ISBN #1}\fi
\ifx \binits  \undefined \def \binits#1{#1}\fi
\ifx \bauthor  \undefined \def \bauthor#1{#1}\fi
\ifx \batitle  \undefined \def \batitle#1{#1}\fi
\ifx \bjtitle  \undefined \def \bjtitle#1{#1}\fi
\ifx \bvolume  \undefined \def \bvolume#1{\textbf{#1}}\fi
\ifx \byear  \undefined \def \byear#1{#1}\fi
\ifx \bissue  \undefined \def \bissue#1{#1}\fi
\ifx \bfpage  \undefined \def \bfpage#1{#1}\fi
\ifx \blpage  \undefined \def \blpage #1{#1}\fi
\ifx \burl  \undefined \def \burl#1{\textsf{#1}}\fi
\ifx \doiurl  \undefined \def \doiurl#1{\url{https://doi.org/#1}}\fi
\ifx \betal  \undefined \def \betal{\textit{et al.}}\fi
\ifx \binstitute  \undefined \def \binstitute#1{#1}\fi
\ifx \binstitutionaled  \undefined \def \binstitutionaled#1{#1}\fi
\ifx \bctitle  \undefined \def \bctitle#1{#1}\fi
\ifx \beditor  \undefined \def \beditor#1{#1}\fi
\ifx \bpublisher  \undefined \def \bpublisher#1{#1}\fi
\ifx \bbtitle  \undefined \def \bbtitle#1{#1}\fi
\ifx \bedition  \undefined \def \bedition#1{#1}\fi
\ifx \bseriesno  \undefined \def \bseriesno#1{#1}\fi
\ifx \blocation  \undefined \def \blocation#1{#1}\fi
\ifx \bsertitle  \undefined \def \bsertitle#1{#1}\fi
\ifx \bsnm \undefined \def \bsnm#1{#1}\fi
\ifx \bsuffix \undefined \def \bsuffix#1{#1}\fi
\ifx \bparticle \undefined \def \bparticle#1{#1}\fi
\ifx \barticle \undefined \def \barticle#1{#1}\fi
\bibcommenthead
\ifx \bconfdate \undefined \def \bconfdate #1{#1}\fi
\ifx \botherref \undefined \def \botherref #1{#1}\fi
\ifx \url \undefined \def \url#1{\textsf{#1}}\fi
\ifx \bchapter \undefined \def \bchapter#1{#1}\fi
\ifx \bbook \undefined \def \bbook#1{#1}\fi
\ifx \bcomment \undefined \def \bcomment#1{#1}\fi
\ifx \oauthor \undefined \def \oauthor#1{#1}\fi
\ifx \citeauthoryear \undefined \def \citeauthoryear#1{#1}\fi
\ifx \endbibitem  \undefined \def \endbibitem {}\fi
\ifx \bconflocation  \undefined \def \bconflocation#1{#1}\fi
\ifx \arxivurl  \undefined \def \arxivurl#1{\textsf{#1}}\fi
\csname PreBibitemsHook\endcsname

%%% 1
\bibitem[\protect\citeauthoryear{Settles}{1995}]{settles1995active}
\begin{barticle}
\bauthor{\bsnm{Settles}, \binits{B.}}:
\batitle{Active learning literature survey}.
\bjtitle{Science}
\bvolume{10}(\bissue{3}),
\bfpage{237}--\blpage{304}
(\byear{1995})
\end{barticle}
\endbibitem

%%% 2
\bibitem[\protect\citeauthoryear{Cai et~al.}{2021}]{segmentation3}
\begin{bchapter}
\bauthor{\bsnm{Cai}, \binits{L.}},
\bauthor{\bsnm{Xu}, \binits{X.}},
\bauthor{\bsnm{Liew}, \binits{J.H.}},
\bauthor{\bsnm{Foo}, \binits{C.S.}}:
\bctitle{Revisiting superpixels for active learning in semantic segmentation
  with realistic annotation costs}.
In: \bbtitle{Proceedings of the IEEE/CVF Conference on Computer Vision and
  Pattern Recognition},
pp. \bfpage{10988}--\blpage{10997}
(\byear{2021})
\end{bchapter}
\endbibitem

%%% 3
\bibitem[\protect\citeauthoryear{Haussmann et~al.}{2020}]{detection2}
\begin{bchapter}
\bauthor{\bsnm{Haussmann}, \binits{E.}},
\bauthor{\bsnm{Fenzi}, \binits{M.}},
\bauthor{\bsnm{Chitta}, \binits{K.}},
\bauthor{\bsnm{Ivanecky}, \binits{J.}},
\bauthor{\bsnm{Xu}, \binits{H.}},
\bauthor{\bsnm{Roy}, \binits{D.}},
\bauthor{\bsnm{Mittel}, \binits{A.}},
\bauthor{\bsnm{Koumchatzky}, \binits{N.}},
\bauthor{\bsnm{Farabet}, \binits{C.}},
\bauthor{\bsnm{Alvarez}, \binits{J.M.}}:
\bctitle{Scalable active learning for object detection}.
In: \bbtitle{2020 IEEE Intelligent Vehicles Symposium},
pp. \bfpage{1430}--\blpage{1435}
(\byear{2020})
\end{bchapter}
\endbibitem

%%% 4
\bibitem[\protect\citeauthoryear{Joshi et~al.}{2009}]{joshi2009multi}
\begin{bchapter}
\bauthor{\bsnm{Joshi}, \binits{A.J.}},
\bauthor{\bsnm{Porikli}, \binits{F.}},
\bauthor{\bsnm{Papanikolopoulos}, \binits{N.}}:
\bctitle{Multi-class active learning for image classification}.
In: \bbtitle{2009 IEEE Conference on Computer Vision and Pattern Recognition},
pp. \bfpage{2372}--\blpage{2379}
(\byear{2009})
\end{bchapter}
\endbibitem

%%% 5
\bibitem[\protect\citeauthoryear{Luo et~al.}{2013}]{luo2013latent}
\begin{bchapter}
\bauthor{\bsnm{Luo}, \binits{W.}},
\bauthor{\bsnm{Schwing}, \binits{A.G.}},
\bauthor{\bsnm{Urtasun}, \binits{R.}}:
\bctitle{Latent structured active learning}.
In: \bbtitle{Proceedings of the 26th International Conference on Neural
  Information Processing Systems},
vol. \bseriesno{1},
pp. \bfpage{728}--\blpage{736}
(\byear{2013})
\end{bchapter}
\endbibitem

%%% 6
\bibitem[\protect\citeauthoryear{Yoo and Kweon}{2019}]{lloss}
\begin{bchapter}
\bauthor{\bsnm{Yoo}, \binits{D.}},
\bauthor{\bsnm{Kweon}, \binits{I.S.}}:
\bctitle{Learning loss for active learning}.
In: \bbtitle{Proceedings of the IEEE/CVF Conference on Computer Vision and
  Pattern Recognition},
pp. \bfpage{93}--\blpage{102}
(\byear{2019})
\end{bchapter}
\endbibitem

%%% 7
\bibitem[\protect\citeauthoryear{Kirsch et~al.}{2019}]{batchbald}
\begin{bchapter}
\bauthor{\bsnm{Kirsch}, \binits{A.}},
\bauthor{\bsnm{Amersfoort}, \binits{J.v.}},
\bauthor{\bsnm{Gal}, \binits{Y.}}:
\bctitle{Batchbald: efficient and diverse batch acquisition for deep bayesian
  active learning}.
In: \bbtitle{Proceedings of the 33rd International Conference on Neural
  Information Processing Systems},
vol. \bseriesno{32},
pp. \bfpage{7026}--\blpage{7037}
(\byear{2019})
\end{bchapter}
\endbibitem

%%% 8
\bibitem[\protect\citeauthoryear{Kim et~al.}{2021}]{tvaal}
\begin{bchapter}
\bauthor{\bsnm{Kim}, \binits{K.}},
\bauthor{\bsnm{Park}, \binits{D.}},
\bauthor{\bsnm{Kim}, \binits{K.I.}},
\bauthor{\bsnm{Chun}, \binits{S.Y.}}:
\bctitle{Task-aware variational adversarial active learning}.
In: \bbtitle{Proceedings of the IEEE/CVF Conference on Computer Vision and
  Pattern Recognition},
pp. \bfpage{8166}--\blpage{8175}
(\byear{2021})
\end{bchapter}
\endbibitem

%%% 9
\bibitem[\protect\citeauthoryear{Parvaneh et~al.}{2022}]{parvaneh2022active}
\begin{bchapter}
\bauthor{\bsnm{Parvaneh}, \binits{A.}},
\bauthor{\bsnm{Abbasnejad}, \binits{E.}},
\bauthor{\bsnm{Teney}, \binits{D.}},
\bauthor{\bsnm{Haffari}, \binits{G.R.}},
\bauthor{\bsnm{Hengel}, \binits{A.}},
\bauthor{\bsnm{Shi}, \binits{J.Q.}}:
\bctitle{Active learning by feature mixing}.
In: \bbtitle{Proceedings of the IEEE/CVF Conference on Computer Vision and
  Pattern Recognition},
pp. \bfpage{12237}--\blpage{12246}
(\byear{2022})
\end{bchapter}
\endbibitem

%%% 10
\bibitem[\protect\citeauthoryear{Fang and Zhu}{2012}]{fang2012don}
\begin{bchapter}
\bauthor{\bsnm{Fang}, \binits{M.}},
\bauthor{\bsnm{Zhu}, \binits{X.}}:
\bctitle{I don't know the label: Active learning with blind knowledge}.
In: \bbtitle{Proceedings of the 21st International Conference on Pattern
  Recognition},
pp. \bfpage{2238}--\blpage{2241}
(\byear{2012})
\end{bchapter}
\endbibitem

%%% 11
\bibitem[\protect\citeauthoryear{Briggs et~al.}{2012}]{briggs2012acoustic}
\begin{barticle}
\bauthor{\bsnm{Briggs}, \binits{F.}},
\bauthor{\bsnm{Lakshminarayanan}, \binits{B.}},
\bauthor{\bsnm{Neal}, \binits{L.}},
\bauthor{\bsnm{Fern}, \binits{X.Z.}},
\bauthor{\bsnm{Raich}, \binits{R.}},
\bauthor{\bsnm{Hadley}, \binits{S.J.}},
\bauthor{\bsnm{Hadley}, \binits{A.S.}},
\bauthor{\bsnm{Betts}, \binits{M.G.}}:
\batitle{Acoustic classification of multiple simultaneous bird species: A
  multi-instance multi-label approach}.
\bjtitle{The Journal of the Acoustical Society of America}
\bvolume{131}(\bissue{6}),
\bfpage{4640}--\blpage{4650}
(\byear{2012})
\end{barticle}
\endbibitem

%%% 12
\bibitem[\protect\citeauthoryear{Cour et~al.}{2011}]{cour2011learning}
\begin{barticle}
\bauthor{\bsnm{Cour}, \binits{T.}},
\bauthor{\bsnm{Sapp}, \binits{B.}},
\bauthor{\bsnm{Taskar}, \binits{B.}}:
\batitle{Learning from partial labels}.
\bjtitle{The Journal of Machine Learning Research}
\bvolume{12},
\bfpage{1501}--\blpage{1536}
(\byear{2011})
\end{barticle}
\endbibitem

%%% 13
\bibitem[\protect\citeauthoryear{Zeng et~al.}{2013}]{zeng2013learning}
\begin{bchapter}
\bauthor{\bsnm{Zeng}, \binits{Z.}},
\bauthor{\bsnm{Xiao}, \binits{S.}},
\bauthor{\bsnm{Jia}, \binits{K.}},
\bauthor{\bsnm{Chan}, \binits{T.-H.}},
\bauthor{\bsnm{Gao}, \binits{S.}},
\bauthor{\bsnm{Xu}, \binits{D.}},
\bauthor{\bsnm{Ma}, \binits{Y.}}:
\bctitle{Learning by associating ambiguously labeled images}.
In: \bbtitle{Proceedings of the IEEE Conference on Computer Vision and Pattern
  Recognition},
pp. \bfpage{708}--\blpage{715}
(\byear{2013})
\end{bchapter}
\endbibitem

%%% 14
\bibitem[\protect\citeauthoryear{Feng and An}{2018}]{feng2018leveraging}
\begin{bchapter}
\bauthor{\bsnm{Feng}, \binits{L.}},
\bauthor{\bsnm{An}, \binits{B.}}:
\bctitle{Leveraging latent label distributions for partial label learning.}
In: \bbtitle{International Joint Conference on Artificial Intelligence},
pp. \bfpage{2107}--\blpage{2113}
(\byear{2018})
\end{bchapter}
\endbibitem

%%% 15
\bibitem[\protect\citeauthoryear{Wang et~al.}{2019}]{wang2019adaptive}
\begin{bchapter}
\bauthor{\bsnm{Wang}, \binits{D.-B.}},
\bauthor{\bsnm{Li}, \binits{L.}},
\bauthor{\bsnm{Zhang}, \binits{M.-L.}}:
\bctitle{Adaptive graph guided disambiguation for partial label learning}.
In: \bbtitle{Proceedings of the 25th ACM SIGKDD International Conference on
  Knowledge Discovery \& Data Mining},
pp. \bfpage{83}--\blpage{91}
(\byear{2019})
\end{bchapter}
\endbibitem

%%% 16
\bibitem[\protect\citeauthoryear{Wang et~al.}{2022}]{wang2022pico}
\begin{bchapter}
\bauthor{\bsnm{Wang}, \binits{H.}},
\bauthor{\bsnm{Xiao}, \binits{R.}},
\bauthor{\bsnm{Li}, \binits{Y.}},
\bauthor{\bsnm{Feng}, \binits{L.}},
\bauthor{\bsnm{Niu}, \binits{G.}},
\bauthor{\bsnm{Chen}, \binits{G.}},
\bauthor{\bsnm{Zhao}, \binits{J.}}:
\bctitle{Pico: Contrastive label disambiguation for partial label learning}.
In: \bbtitle{International Conference on Learning Representations}
(\byear{2022})
\end{bchapter}
\endbibitem

%%% 17
\bibitem[\protect\citeauthoryear{Zhang et~al.}{2022}]{zhang2022exploiting}
\begin{bchapter}
\bauthor{\bsnm{Zhang}, \binits{F.}},
\bauthor{\bsnm{Feng}, \binits{L.}},
\bauthor{\bsnm{Han}, \binits{B.}},
\bauthor{\bsnm{Liu}, \binits{T.}},
\bauthor{\bsnm{Niu}, \binits{G.}},
\bauthor{\bsnm{Qin}, \binits{T.}},
\bauthor{\bsnm{Sugiyama}, \binits{M.}}:
\bctitle{Exploiting class activation value for partial-label learning}.
In: \bbtitle{International Conference on Learning Representations}
(\byear{2022})
\end{bchapter}
\endbibitem

%%% 18
\bibitem[\protect\citeauthoryear{Feng et~al.}{2020}]{feng2020provably}
\begin{bchapter}
\bauthor{\bsnm{Feng}, \binits{L.}},
\bauthor{\bsnm{Lv}, \binits{J.}},
\bauthor{\bsnm{Han}, \binits{B.}},
\bauthor{\bsnm{Xu}, \binits{M.}},
\bauthor{\bsnm{Niu}, \binits{G.}},
\bauthor{\bsnm{Geng}, \binits{X.}},
\bauthor{\bsnm{An}, \binits{B.}},
\bauthor{\bsnm{Sugiyama}, \binits{M.}}:
\bctitle{Provably consistent partial-label learning}.
In: \bbtitle{Advances in Neural Information Processing Systems}
(\byear{2020})
\end{bchapter}
\endbibitem

%%% 19
\bibitem[\protect\citeauthoryear{Lv et~al.}{2020}]{lv2020progressive}
\begin{bchapter}
\bauthor{\bsnm{Lv}, \binits{J.}},
\bauthor{\bsnm{Xu}, \binits{M.}},
\bauthor{\bsnm{Feng}, \binits{L.}},
\bauthor{\bsnm{Niu}, \binits{G.}},
\bauthor{\bsnm{Geng}, \binits{X.}},
\bauthor{\bsnm{Sugiyama}, \binits{M.}}:
\bctitle{Progressive identification of true labels for partial-label learning}.
In: \bbtitle{International Conference on Machine Learning},
pp. \bfpage{6500}--\blpage{6510}
(\byear{2020})
\end{bchapter}
\endbibitem

%%% 20
\bibitem[\protect\citeauthoryear{Wen et~al.}{2021}]{wen2021leveraged}
\begin{bchapter}
\bauthor{\bsnm{Wen}, \binits{H.}},
\bauthor{\bsnm{Cui}, \binits{J.}},
\bauthor{\bsnm{Hang}, \binits{H.}},
\bauthor{\bsnm{Liu}, \binits{J.}},
\bauthor{\bsnm{Wang}, \binits{Y.}},
\bauthor{\bsnm{Lin}, \binits{Z.}}:
\bctitle{Leveraged weighted loss for partial label learning}.
In: \bbtitle{Proceedings of the 38th International Conference on Machine
  Learning},
vol. \bseriesno{139},
pp. \bfpage{11091}--\blpage{11100}
(\byear{2021})
\end{bchapter}
\endbibitem

%%% 21
\bibitem[\protect\citeauthoryear{Chen et~al.}{2006}]{chen2006empirical}
\begin{bchapter}
\bauthor{\bsnm{Chen}, \binits{J.}},
\bauthor{\bsnm{Schein}, \binits{A.}},
\bauthor{\bsnm{Ungar}, \binits{L.}},
\bauthor{\bsnm{Palmer}, \binits{M.}}:
\bctitle{An empirical study of the behavior of active learning for word sense
  disambiguation}.
In: \bbtitle{Proceedings of the Human Language Technology Conference of the
  NAACL, Main Conference},
pp. \bfpage{120}--\blpage{127}
(\byear{2006})
\end{bchapter}
\endbibitem

%%% 22
\bibitem[\protect\citeauthoryear{Perez and Wang}{2017}]{perez2017effectiveness}
\begin{botherref}
\oauthor{\bsnm{Perez}, \binits{L.}},
\oauthor{\bsnm{Wang}, \binits{J.}}:
The effectiveness of data augmentation in image classification using deep
  learning
(2017).
Preprint at \url{https://arxiv.org/abs/1712.04621 }
\end{botherref}
\endbibitem

%%% 23
\bibitem[\protect\citeauthoryear{Shorten and
  Khoshgoftaar}{2019}]{shorten2019survey}
\begin{barticle}
\bauthor{\bsnm{Shorten}, \binits{C.}},
\bauthor{\bsnm{Khoshgoftaar}, \binits{T.M.}}:
\batitle{A survey on image data augmentation for deep learning}.
\bjtitle{Journal of big data}
\bvolume{6}(\bissue{1}),
\bfpage{1}--\blpage{48}
(\byear{2019})
\end{barticle}
\endbibitem

%%% 24
\bibitem[\protect\citeauthoryear{De~Neys et~al.}{2005}]{mental1}
\begin{barticle}
\bauthor{\bsnm{De~Neys}, \binits{W.}},
\bauthor{\bsnm{Schaeken}, \binits{W.}},
\bauthor{\bsnm{d'Ydewalle}, \binits{G.}}:
\batitle{Working memory and everyday conditional reasoning: Retrieval and
  inhibition of stored counterexamples}.
\bjtitle{Thinking \& Reasoning}
\bvolume{11}(\bissue{4}),
\bfpage{349}--\blpage{381}
(\byear{2005})
\end{barticle}
\endbibitem

%%% 25
\bibitem[\protect\citeauthoryear{Verschueren et~al.}{2005}]{mental2}
\begin{barticle}
\bauthor{\bsnm{Verschueren}, \binits{N.}},
\bauthor{\bsnm{Schaeken}, \binits{W.}},
\bauthor{\bsnm{d’Ydewalle}, \binits{G.}}:
\batitle{Everyday conditional reasoning: A working memory—dependent tradeoff
  between counterexample and likelihood use}.
\bjtitle{Memory \& Cognition}
\bvolume{33}(\bissue{1}),
\bfpage{107}--\blpage{119}
(\byear{2005})
\end{barticle}
\endbibitem

%%% 26
\bibitem[\protect\citeauthoryear{Johnson-Laird}{2010}]{mental3}
\begin{barticle}
\bauthor{\bsnm{Johnson-Laird}, \binits{P.N.}}:
\batitle{Mental models and human reasoning}.
\bjtitle{Proceedings of the National Academy of Sciences}
\bvolume{107}(\bissue{43}),
\bfpage{18243}--\blpage{18250}
(\byear{2010})
\end{barticle}
\endbibitem

%%% 27
\bibitem[\protect\citeauthoryear{Lewis and
  Catlett}{1994}]{lewis1994heterogenous}
\begin{bchapter}
\bauthor{\bsnm{Lewis}, \binits{D.D.}},
\bauthor{\bsnm{Catlett}, \binits{J.}}:
\bctitle{Heterogenous uncertainty sampling for supervised learning}.
In: \bbtitle{Proceedings of the Eleventh International Conference on
  International Conference on Machine Learning},
pp. \bfpage{148}--\blpage{156}
(\byear{1994})
\end{bchapter}
\endbibitem

%%% 28
\bibitem[\protect\citeauthoryear{Roth and Small}{2006}]{roth2006margin}
\begin{bchapter}
\bauthor{\bsnm{Roth}, \binits{D.}},
\bauthor{\bsnm{Small}, \binits{K.}}:
\bctitle{Margin-based active learning for structured output spaces}.
In: \bbtitle{European Conference on Machine Learning},
pp. \bfpage{413}--\blpage{424}
(\byear{2006})
\end{bchapter}
\endbibitem

%%% 29
\bibitem[\protect\citeauthoryear{Gal et~al.}{2017}]{bald}
\begin{bchapter}
\bauthor{\bsnm{Gal}, \binits{Y.}},
\bauthor{\bsnm{Islam}, \binits{R.}},
\bauthor{\bsnm{Ghahramani}, \binits{Z.}}:
\bctitle{Deep bayesian active learning with image data}.
In: \bbtitle{International Conference on Machine Learning},
pp. \bfpage{1183}--\blpage{1192}
(\byear{2017})
\end{bchapter}
\endbibitem

%%% 30
\bibitem[\protect\citeauthoryear{Settles et~al.}{2007}]{settles2007multiple}
\begin{botherref}
\oauthor{\bsnm{Settles}, \binits{B.}},
\oauthor{\bsnm{Craven}, \binits{M.}},
\oauthor{\bsnm{Ray}, \binits{S.}}:
Multiple-instance active learning.
Advances in neural information processing systems
\textbf{20}
(2007)
\end{botherref}
\endbibitem

%%% 31
\bibitem[\protect\citeauthoryear{Ash et~al.}{2020}]{badge}
\begin{bchapter}
\bauthor{\bsnm{Ash}, \binits{J.T.}},
\bauthor{\bsnm{Zhang}, \binits{C.}},
\bauthor{\bsnm{Krishnamurthy}, \binits{A.}},
\bauthor{\bsnm{Langford}, \binits{J.}},
\bauthor{\bsnm{Agarwal}, \binits{A.}}:
\bctitle{Deep batch active learning by diverse, uncertain gradient lower
  bounds}.
In: \bbtitle{International Conference on Learning Representations}
(\byear{2020})
\end{bchapter}
\endbibitem

%%% 32
\bibitem[\protect\citeauthoryear{Elhamifar et~al.}{2013}]{elhamifar2013convex}
\begin{bchapter}
\bauthor{\bsnm{Elhamifar}, \binits{E.}},
\bauthor{\bsnm{Sapiro}, \binits{G.}},
\bauthor{\bsnm{Yang}, \binits{A.}},
\bauthor{\bsnm{Sasrty}, \binits{S.S.}}:
\bctitle{A convex optimization framework for active learning}.
In: \bbtitle{Proceedings of the IEEE International Conference on Computer
  Vision},
pp. \bfpage{209}--\blpage{216}
(\byear{2013})
\end{bchapter}
\endbibitem

%%% 33
\bibitem[\protect\citeauthoryear{Yang et~al.}{2015}]{yang2015multi}
\begin{barticle}
\bauthor{\bsnm{Yang}, \binits{Y.}},
\bauthor{\bsnm{Ma}, \binits{Z.}},
\bauthor{\bsnm{Nie}, \binits{F.}},
\bauthor{\bsnm{Chang}, \binits{X.}},
\bauthor{\bsnm{Hauptmann}, \binits{A.G.}}:
\batitle{Multi-class active learning by uncertainty sampling with diversity
  maximization}.
\bjtitle{International Journal of Computer Vision}
\bvolume{113}(\bissue{2}),
\bfpage{113}--\blpage{127}
(\byear{2015})
\end{barticle}
\endbibitem

%%% 34
\bibitem[\protect\citeauthoryear{Nguyen and Smeulders}{2004}]{nguyen2004active}
\begin{bchapter}
\bauthor{\bsnm{Nguyen}, \binits{H.T.}},
\bauthor{\bsnm{Smeulders}, \binits{A.}}:
\bctitle{Active learning using pre-clustering}.
In: \bbtitle{Proceedings of the Twenty-first International Conference on
  Machine Learning},
p. \bfpage{79}
(\byear{2004})
\end{bchapter}
\endbibitem

%%% 35
\bibitem[\protect\citeauthoryear{Freytag et~al.}{2014}]{freytag2014selecting}
\begin{bchapter}
\bauthor{\bsnm{Freytag}, \binits{A.}},
\bauthor{\bsnm{Rodner}, \binits{E.}},
\bauthor{\bsnm{Denzler}, \binits{J.}}:
\bctitle{Selecting influential examples: Active learning with expected model
  output changes}.
In: \bbtitle{European Conference on Computer Vision},
pp. \bfpage{562}--\blpage{577}
(\byear{2014}).
\bcomment{Springer}
\end{bchapter}
\endbibitem

%%% 36
\bibitem[\protect\citeauthoryear{Sener and Savarese}{2018}]{coreset}
\begin{bchapter}
\bauthor{\bsnm{Sener}, \binits{O.}},
\bauthor{\bsnm{Savarese}, \binits{S.}}:
\bctitle{Active learning for convolutional neural networks: A core-set
  approach}.
In: \bbtitle{International Conference on Learning Representations}
(\byear{2018})
\end{bchapter}
\endbibitem

%%% 37
\bibitem[\protect\citeauthoryear{Sinha et~al.}{2019}]{vaal}
\begin{bchapter}
\bauthor{\bsnm{Sinha}, \binits{S.}},
\bauthor{\bsnm{Ebrahimi}, \binits{S.}},
\bauthor{\bsnm{Darrell}, \binits{T.}}:
\bctitle{Variational adversarial active learning}.
In: \bbtitle{Proceedings of the IEEE/CVF International Conference on Computer
  Vision},
pp. \bfpage{5972}--\blpage{5981}
(\byear{2019})
\end{bchapter}
\endbibitem

%%% 38
\bibitem[\protect\citeauthoryear{Donmez and
  Carbonell}{2008}]{donmez2008proactive}
\begin{bchapter}
\bauthor{\bsnm{Donmez}, \binits{P.}},
\bauthor{\bsnm{Carbonell}, \binits{J.G.}}:
\bctitle{Proactive learning: cost-sensitive active learning with multiple
  imperfect oracles}.
In: \bbtitle{Proceedings of the 17th ACM Conference on Information and
  Knowledge Management},
pp. \bfpage{619}--\blpage{628}
(\byear{2008})
\end{bchapter}
\endbibitem

%%% 39
\bibitem[\protect\citeauthoryear{Du and Ling}{2010}]{du2010active}
\begin{bchapter}
\bauthor{\bsnm{Du}, \binits{J.}},
\bauthor{\bsnm{Ling}, \binits{C.X.}}:
\bctitle{Active learning with human-like noisy oracle}.
In: \bbtitle{2010 IEEE International Conference on Data Mining},
pp. \bfpage{797}--\blpage{802}
(\byear{2010})
\end{bchapter}
\endbibitem

%%% 40
\bibitem[\protect\citeauthoryear{Yan et~al.}{2016}]{yan2016active}
\begin{botherref}
\oauthor{\bsnm{Yan}, \binits{S.}},
\oauthor{\bsnm{Chaudhuri}, \binits{K.}},
\oauthor{\bsnm{Javidi}, \binits{T.}}:
Active learning from imperfect labelers.
Advances in Neural Information Processing Systems
\textbf{29}
(2016)
\end{botherref}
\endbibitem

%%% 41
\bibitem[\protect\citeauthoryear{Chakraborty}{2020}]{chakraborty2020asking}
\begin{bchapter}
\bauthor{\bsnm{Chakraborty}, \binits{S.}}:
\bctitle{Asking the right questions to the right users: Active learning with
  imperfect oracles}.
In: \bbtitle{Proceedings of the AAAI Conference on Artificial Intelligence},
vol. \bseriesno{34},
pp. \bfpage{3365}--\blpage{3372}
(\byear{2020})
\end{bchapter}
\endbibitem

%%% 42
\bibitem[\protect\citeauthoryear{H{\"u}llermeier and
  Beringer}{2006}]{hullermeier2006learning}
\begin{barticle}
\bauthor{\bsnm{H{\"u}llermeier}, \binits{E.}},
\bauthor{\bsnm{Beringer}, \binits{J.}}:
\batitle{Learning from ambiguously labeled examples}.
\bjtitle{Intelligent Data Analysis}
\bvolume{10}(\bissue{5}),
\bfpage{419}--\blpage{439}
(\byear{2006})
\end{barticle}
\endbibitem

%%% 43
\bibitem[\protect\citeauthoryear{Gong et~al.}{2017}]{gong2017regularization}
\begin{barticle}
\bauthor{\bsnm{Gong}, \binits{C.}},
\bauthor{\bsnm{Liu}, \binits{T.}},
\bauthor{\bsnm{Tang}, \binits{Y.}},
\bauthor{\bsnm{Yang}, \binits{J.}},
\bauthor{\bsnm{Yang}, \binits{J.}},
\bauthor{\bsnm{Tao}, \binits{D.}}:
\batitle{A regularization approach for instance-based superset label learning}.
\bjtitle{IEEE Transactions on Cybernetics}
\bvolume{48}(\bissue{3}),
\bfpage{967}--\blpage{978}
(\byear{2017})
\end{barticle}
\endbibitem

%%% 44
\bibitem[\protect\citeauthoryear{Cour et~al.}{2009}]{cour2009learning}
\begin{bchapter}
\bauthor{\bsnm{Cour}, \binits{T.}},
\bauthor{\bsnm{Sapp}, \binits{B.}},
\bauthor{\bsnm{Jordan}, \binits{C.}},
\bauthor{\bsnm{Taskar}, \binits{B.}}:
\bctitle{Learning from ambiguously labeled images}.
In: \bbtitle{2009 IEEE Conference on Computer Vision and Pattern Recognition},
pp. \bfpage{919}--\blpage{926}
(\byear{2009})
\end{bchapter}
\endbibitem

%%% 45
\bibitem[\protect\citeauthoryear{Yao et~al.}{2020}]{yao2020deep}
\begin{bchapter}
\bauthor{\bsnm{Yao}, \binits{Y.}},
\bauthor{\bsnm{Deng}, \binits{J.}},
\bauthor{\bsnm{Chen}, \binits{X.}},
\bauthor{\bsnm{Gong}, \binits{C.}},
\bauthor{\bsnm{Wu}, \binits{J.}},
\bauthor{\bsnm{Yang}, \binits{J.}}:
\bctitle{Deep discriminative cnn with temporal ensembling for
  ambiguously-labeled image classification}.
In: \bbtitle{Proceedings of the AAAI Conference on Artificial Intelligence},
vol. \bseriesno{34},
pp. \bfpage{12669}--\blpage{12676}
(\byear{2020})
\end{bchapter}
\endbibitem

%%% 46
\bibitem[\protect\citeauthoryear{Liu and
  Dietterich}{2014}]{liu2014learnability}
\begin{bchapter}
\bauthor{\bsnm{Liu}, \binits{L.}},
\bauthor{\bsnm{Dietterich}, \binits{T.}}:
\bctitle{Learnability of the superset label learning problem}.
In: \bbtitle{International Conference on Machine Learning},
pp. \bfpage{1629}--\blpage{1637}
(\byear{2014})
\end{bchapter}
\endbibitem

%%% 47
\bibitem[\protect\citeauthoryear{Yu and Zhang}{2016}]{yu2016maximum}
\begin{bchapter}
\bauthor{\bsnm{Yu}, \binits{F.}},
\bauthor{\bsnm{Zhang}, \binits{M.-L.}}:
\bctitle{Maximum margin partial label learning}.
In: \bbtitle{Asian Conference on Machine Learning},
pp. \bfpage{96}--\blpage{111}
(\byear{2016})
\end{bchapter}
\endbibitem

%%% 48
\bibitem[\protect\citeauthoryear{Feng and An}{2019}]{feng2019partial}
\begin{bchapter}
\bauthor{\bsnm{Feng}, \binits{L.}},
\bauthor{\bsnm{An}, \binits{B.}}:
\bctitle{Partial label learning by semantic difference maximization.}
In: \bbtitle{International Joint Conference on Artificial Intelligence},
pp. \bfpage{2294}--\blpage{2300}
(\byear{2019})
\end{bchapter}
\endbibitem

%%% 49
\bibitem[\protect\citeauthoryear{Yan and Guo}{2020}]{yan2020partial}
\begin{bchapter}
\bauthor{\bsnm{Yan}, \binits{Y.}},
\bauthor{\bsnm{Guo}, \binits{Y.}}:
\bctitle{Partial label learning with batch label correction}.
In: \bbtitle{Proceedings of the AAAI Conference on Artificial Intelligence},
vol. \bseriesno{34},
pp. \bfpage{6575}--\blpage{6582}
(\byear{2020})
\end{bchapter}
\endbibitem

%%% 50
\bibitem[\protect\citeauthoryear{Tran et~al.}{2019}]{tran2019bayesian}
\begin{bchapter}
\bauthor{\bsnm{Tran}, \binits{T.}},
\bauthor{\bsnm{Do}, \binits{T.-T.}},
\bauthor{\bsnm{Reid}, \binits{I.}},
\bauthor{\bsnm{Carneiro}, \binits{G.}}:
\bctitle{Bayesian generative active deep learning}.
In: \bbtitle{International Conference on Machine Learning},
pp. \bfpage{6295}--\blpage{6304}
(\byear{2019})
\end{bchapter}
\endbibitem

%%% 51
\bibitem[\protect\citeauthoryear{Ishida et~al.}{2017}]{ishida2017learning}
\begin{botherref}
\oauthor{\bsnm{Ishida}, \binits{T.}},
\oauthor{\bsnm{Niu}, \binits{G.}},
\oauthor{\bsnm{Hu}, \binits{W.}},
\oauthor{\bsnm{Sugiyama}, \binits{M.}}:
Learning from complementary labels.
Advances in neural information processing systems
\textbf{30}
(2017)
\end{botherref}
\endbibitem

%%% 52
\bibitem[\protect\citeauthoryear{LeCun et~al.}{1998}]{minist}
\begin{barticle}
\bauthor{\bsnm{LeCun}, \binits{Y.}},
\bauthor{\bsnm{Bottou}, \binits{L.}},
\bauthor{\bsnm{Bengio}, \binits{Y.}},
\bauthor{\bsnm{Haffner}, \binits{P.}}:
\batitle{Gradient-based learning applied to document recognition}.
\bjtitle{Proceedings of the IEEE}
\bvolume{86}(\bissue{11}),
\bfpage{2278}--\blpage{2324}
(\byear{1998})
\end{barticle}
\endbibitem

%%% 53
\bibitem[\protect\citeauthoryear{Xiao et~al.}{2017}]{fashion}
\begin{botherref}
\oauthor{\bsnm{Xiao}, \binits{H.}},
\oauthor{\bsnm{Rasul}, \binits{K.}},
\oauthor{\bsnm{Vollgraf}, \binits{R.}}:
Fashion-mnist: a novel image dataset for benchmarking machine learning
  algorithms
(2017).
Preprint at \url{https://arxiv.org/abs/1708.07747 }
\end{botherref}
\endbibitem

%%% 54
\bibitem[\protect\citeauthoryear{Netzer et~al.}{2011}]{svhn}
\begin{botherref}
\oauthor{\bsnm{Netzer}, \binits{Y.}},
\oauthor{\bsnm{Wang}, \binits{T.}},
\oauthor{\bsnm{Coates}, \binits{A.}},
\oauthor{\bsnm{Bissacco}, \binits{A.}},
\oauthor{\bsnm{Wu}, \binits{B.}},
\oauthor{\bsnm{Ng}, \binits{A.Y.}}:
Reading digits in natural images with unsupervised feature learning
(2011)
\end{botherref}
\endbibitem

%%% 55
\bibitem[\protect\citeauthoryear{Krizhevsky et~al.}{2009}]{cifar}
\begin{botherref}
\oauthor{\bsnm{Krizhevsky}, \binits{A.}},
\oauthor{\bsnm{Hinton}, \binits{G.}}, et al.:
Learning multiple layers of features from tiny images
(2009)
\end{botherref}
\endbibitem

%%% 56
\bibitem[\protect\citeauthoryear{He et~al.}{2016}]{resnet}
\begin{bchapter}
\bauthor{\bsnm{He}, \binits{K.}},
\bauthor{\bsnm{Zhang}, \binits{X.}},
\bauthor{\bsnm{Ren}, \binits{S.}},
\bauthor{\bsnm{Sun}, \binits{J.}}:
\bctitle{Deep residual learning for image recognition}.
In: \bbtitle{Proceedings of the IEEE Conference on Computer Vision and Pattern
  Recognition},
pp. \bfpage{770}--\blpage{778}
(\byear{2016})
\end{bchapter}
\endbibitem

%%% 57
\bibitem[\protect\citeauthoryear{Simonyan and Zisserman}{2014}]{vgg}
\begin{botherref}
\oauthor{\bsnm{Simonyan}, \binits{K.}},
\oauthor{\bsnm{Zisserman}, \binits{A.}}:
Very deep convolutional networks for large-scale image recognition
(2014).
Preprint at \url{https://arxiv.org/abs/1409.1556 }
\end{botherref}
\endbibitem

%%% 58
\bibitem[\protect\citeauthoryear{Kingma and Ba}{2015}]{adam}
\begin{botherref}
\oauthor{\bsnm{Kingma}, \binits{D.P.}},
\oauthor{\bsnm{Ba}, \binits{J.}}:
Adam: A method for stochastic optimization.
International Conference on Learning Representations
(2015)
\end{botherref}
\endbibitem

%%% 59
\bibitem[\protect\citeauthoryear{Kim et~al.}{2021}]{lada}
\begin{barticle}
\bauthor{\bsnm{Kim}, \binits{Y.-Y.}},
\bauthor{\bsnm{Song}, \binits{K.}},
\bauthor{\bsnm{Jang}, \binits{J.}},
\bauthor{\bsnm{Moon}, \binits{I.-C.}}:
\batitle{Lada: Look-ahead data acquisition via augmentation for deep active
  learning}.
\bjtitle{Advances in Neural Information Processing Systems}
\bvolume{34},
\bfpage{22919}--\blpage{22930}
(\byear{2021})
\end{barticle}
\endbibitem

%%% 60
\bibitem[\protect\citeauthoryear{Liu and Dietterich}{2012}]{liu2012conditional}
\begin{bchapter}
\bauthor{\bsnm{Liu}, \binits{L.}},
\bauthor{\bsnm{Dietterich}, \binits{T.G.}}:
\bctitle{A conditional multinomial mixture model for superset label learning}.
In: \bbtitle{Advances in Neural Information Processing Systems},
pp. \bfpage{548}--\blpage{556}
(\byear{2012})
\end{bchapter}
\endbibitem

%%% 61
\bibitem[\protect\citeauthoryear{Briggs et~al.}{2012}]{briggs2012rank}
\begin{bchapter}
\bauthor{\bsnm{Briggs}, \binits{F.}},
\bauthor{\bsnm{Fern}, \binits{X.Z.}},
\bauthor{\bsnm{Raich}, \binits{R.}}:
\bctitle{Rank-loss support instance machines for miml instance annotation}.
In: \bbtitle{Proceedings of the 18th ACM SIGKDD International Conference on
  Knowledge Discovery and Data Mining},
pp. \bfpage{534}--\blpage{542}
(\byear{2012})
\end{bchapter}
\endbibitem

%%% 62
\bibitem[\protect\citeauthoryear{Guillaumin
  et~al.}{2010}]{guillaumin2010multiple}
\begin{bchapter}
\bauthor{\bsnm{Guillaumin}, \binits{M.}},
\bauthor{\bsnm{Verbeek}, \binits{J.}},
\bauthor{\bsnm{Schmid}, \binits{C.}}:
\bctitle{Multiple instance metric learning from automatically labeled bags of
  faces}.
In: \bbtitle{European Conference on Computer Vision},
pp. \bfpage{634}--\blpage{647}
(\byear{2010})
\end{bchapter}
\endbibitem

%%% 63
\bibitem[\protect\citeauthoryear{Feng and An}{2019}]{self-training}
\begin{bchapter}
\bauthor{\bsnm{Feng}, \binits{L.}},
\bauthor{\bsnm{An}, \binits{B.}}:
\bctitle{Partial label learning with self-guided retraining}.
In: \bbtitle{Proceedings of the AAAI Conference on Artificial Intelligence},
pp. \bfpage{3542}--\blpage{3549}
(\byear{2019})
\end{bchapter}
\endbibitem

\end{thebibliography}

\end{document}